\documentclass[10pt,journal,compsoc]{IEEEtran}

\usepackage{url}
\usepackage{bigints}
\usepackage{amssymb,amstext,amsmath,eqnarray, mathrsfs, bm}
\usepackage{textcomp}
\usepackage{amsthm}
\usepackage{graphicx,caption,subcaption}
\usepackage{siunitx}
\usepackage{multirow,tabularx}
\usepackage{textcomp}
\usepackage{xfrac}

\usepackage{fancyhdr}
\usepackage{algorithm,algpseudocode}
\algnewcommand{\Inputs}[1]{%
  \State \textbf{Inputs:}
  \Statex \hspace*{\algorithmicindent}\parbox[t]{.8\linewidth}{\raggedright #1}
}
\algnewcommand{\Initialize}[1]{%
  \State \textbf{Initialize:}
  \Statex \hspace*{\algorithmicindent}\parbox[t]{.8\linewidth}{\raggedright #1}
}

\def\bg#1{\mbox{\boldmath$#1$}} 
\newtheorem{theorem}{Theorem}

\usepackage[symbol*]{footmisc}
\DefineFNsymbolsTM{myfnsymbols}{
  \textasteriskcentered *
  \textdagger    \dagger
  \textdaggerdbl \ddagger
  \textsection   \mathsection
  \textbardbl    \|%
  \textparagraph \mathparagraph
}%

\ifCLASSOPTIONcompsoc
  \usepackage[nocompress]{cite}
\else
  \usepackage{cite}
\fi

\ifCLASSINFOpdf
\else
\fi

\begin{document}


\chead{IEEE Transactions on Pattern Analysis and Machine Intelligence. (c) 2017 IEEE. DOI: 10.1109/TPAMI.2017.2772235. Author's accepted version. The final publication is available at http://ieeexplore.ieee.org/document/8103808/}

\title{Linear Maximum Margin Classifier for Learning from Uncertain Data}

\author{Christos~Tzelepis,~\IEEEmembership{Student Member,~IEEE, }
	    Vasileios~Mezaris,~\IEEEmembership{Senior Member,~IEEE,  }
	    and~Ioannis~Patras,~\IEEEmembership{Senior Member,~IEEE  }
\thanks{C. Tzelepis is with the School of Electronic Engineering and Computer Science, Queen Mary University of London, London E1 4NS, U.K. and also with the Information Technologies Institute/Centre for Research and Technology Hellas (CERTH), Thermi 57001, Greece, (email: c.tzelepis@qmul.ac.uk).}
\thanks{V. Mezaris is with the Information Technologies Institute/Centre for Research and Technology Hellas (CERTH), Thermi 57001, Greece (email: bmezaris@iti.gr).}
\thanks{I. Patras is with the School of Electronic Engineering and Computer Science, Queen Mary University of London, London E1 4NS, U.K. (e-mail: i.patras@qmul.ac.uk).}}


\IEEEaftertitletext{\vspace{-1\baselineskip}\noindent%
\begin{abstract}
	In this paper, we propose a maximum margin classifier that deals with uncertainty in data input. More specifically, we reformulate the SVM framework such that each training example can be modeled by a multi-dimensional Gaussian distribution described by its mean vector and its covariance matrix -- the latter modeling the uncertainty. We address the classification problem and define a cost function that is the expected value of the classical SVM cost when data samples are drawn from the multi-dimensional Gaussian distributions that form the set of the training examples. Our formulation approximates the classical SVM formulation when the training examples are isotropic Gaussians with variance tending to zero. We arrive at a convex optimization problem, which we solve efficiently in the primal form using a stochastic gradient descent approach. The resulting classifier, which we name SVM with Gaussian Sample Uncertainty (SVM-GSU), is tested on synthetic data and five publicly available and popular datasets; namely, the MNIST, WDBC, DEAP, TV News Channel Commercial Detection, and TRECVID MED datasets. Experimental results verify the effectiveness of the proposed method.
\end{abstract}

\begin{IEEEkeywords}
	Classification, convex optimization, Gaussian anisotropic uncertainty, large margin methods, learning with uncertainty, statistical learning theory
\end{IEEEkeywords}
\vspace{2\baselineskip}}

\maketitle

\IEEEraisesectionheading{\section{Introduction}}

	\IEEEPARstart{S}{upport} Vector Machine (SVM) has been shown to be a powerful paradigm for pattern classification. Its origins can be traced back to~\cite{vladimir1995nature}. Vapnik established the standard regularized SVM algorithm for computing a linear discriminative function that optimizes the margin between the so called support vectors and the separating hyperplane. Despite the fact that the standard SVM algorithm is a well-studied and general framework for statistical learning analysis, it is still an active research field (e.g.,~\cite{solera2016pami, sentelleTNNLS16}).

	However, the classical SVM formulation, as well as the majority of classification methods, do not explicitly model input uncertainty. In standard SVM, each training datum is a vector, whose position in the feature space is considered certain. This does not model the fact that measurement inaccuracies or artifacts of the feature extraction process contaminate the training examples with noise. In several cases the noise distribution is known or can be modeled; e.g., there are cases where each training example represents the average of several measurements or of several samples whose distribution around the mean can be modeled or estimated. Finally, in some cases it is possible to model the process by which the data is generated, for example by modeling the process by which new data is generated from transforms applied on an already given training dataset.

	\begin{figure}[t!]
		\centering
		\includegraphics[width=65mm]{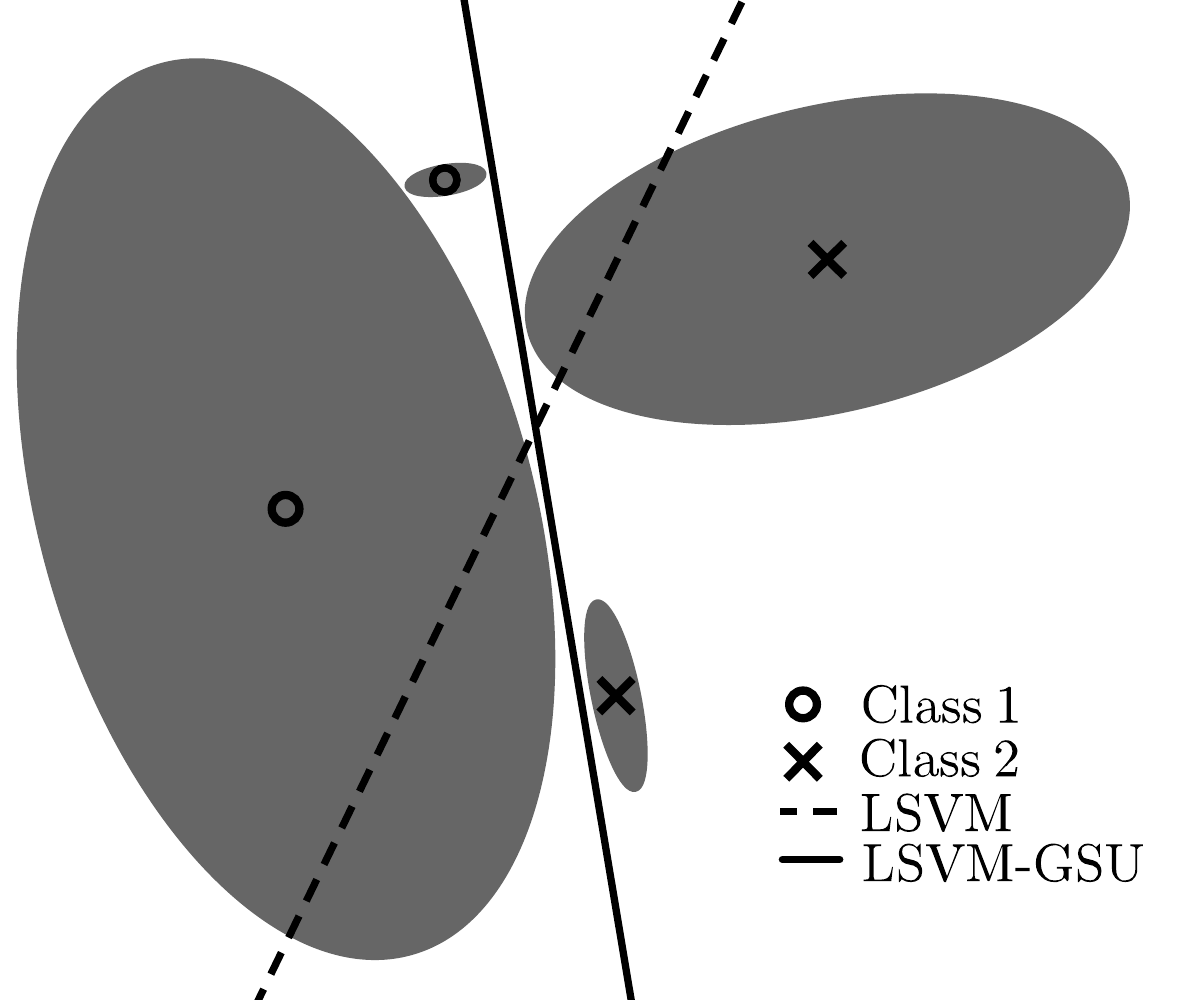}
		\caption{Linear SVM with Gaussian Sample Uncertainty (LSVM-GSU). The solid line depicts the decision boundary of the proposed algorithm, and the dashed line depicts the decision boundary of the standard linear SVM (LSVM).}
		\label{fig:svm_gsu}
	\end{figure}
	    
	In this work, we consider that our training examples are multivariate Gaussian distributions with known means and covariance matrices -- each example having a different covariance matrix expressing the uncertainty around its mean. An illustration is given in Fig.~\ref{fig:svm_gsu}, where the shaded regions are bounded by iso-density loci of the Gaussians, and the means of the Gaussians for examples of the positive and negative classes are located at $\times$ and $\circ$ respectively. A classical SVM formulation would consider only the means of the Gaussians as training examples and, by optimizing the soft margin using the hinge loss and a regularization term, would arrive at the separating hyperplane depicted by the dashed line. In our formulation, we optimize for the soft margin using the same regularization but the {\em expected} value of the hinge loss, where the expectation is taken under the given Gaussians. By doing so, we take into consideration the various uncertainties and arrive at a drastically different decision border, depicted by the solid line in Fig.~\ref{fig:svm_gsu}. It is worth noting that one would arrive at the same decision border with the classical SVM trained on a dataset containing samples drawn from the Gaussians in question, as the number of samples tend to infinity. In addition, our method degenerates to a classical SVM in the case that all of the Gaussians are isotropic with a variance that tends to zero.

	Our work differs from previous works that model uncertainty in the SVM framework either by considering isotropic noise or by using expensive sampling schemes to approximate their loss functions. By contrast, our formulation allows for full covariance matrices that can be different for each example. This allows dealing, among others, with cases where the uncertainty of only a few examples, and/or the uncertainty along only a few of their dimensions, is known or modeled. In the experimental results section we show several real-world problems in which such modeling is beneficial. More specifically, we show cases, in which the variances along (some) of the dimensions are part of the dataset -- this includes medical data where both the means and the variances of several measurements are reported, and large scale video datasets, where the means and the variances of some of the features that are extracted at several time instances in the video in question are reported.  We then show a case in which means and variances are a by-product of the feature extraction method, namely the Welch method for extracting periodograms from temporal EEG data. And finally, we show a case in which, for an image dataset (MNIST) we model the distribution of images under small geometric transforms as Gaussians, using a first-order Taylor approximation to arrive in an analytic form.
	    
	In general, modeling of the uncertainty is a domain- and/or dataset-specific problem, and in this respect, similarly to all of the other methods in the literature that model/use uncertainties, we do not offer a definitive answer on how this can or should be done on any existing dataset. We note, however, that means and (co)-variances are the most commonly reported statistics and that the modeling used in Sect.~\ref{ssec:exp_mnist}, \ref{ssec:exp_deap} could be used also in other similar datasets. In particular, the Taylor expansion method (Appendix~\ref{app:app_B}) that is behind the modeling used in Sect.~\ref{ssec:exp_mnist}, has been used to model the propagation of uncertainties due to a feature extraction process in other domains; for instance, in~\cite{diplaros2006combining} (Sect.~II.B) this is used to model as Gaussian the uncertainty in the estimation of illumination invariant image derivatives. Finally, in our work the cost function, which is based on the expectation of the hinge loss, and its derivatives, can be calculated in closed forms. This allows an efficient implementation using a stochastic gradient descent (SGD) algorithm.
	
	The remainder of this paper is organized as follows. In Section~\ref{sec:rel_work}, we review related work, focusing on SVM-based formulations that explicitly model data uncertainty. In Section~\ref{sec:prop_approach}, we present the proposed algorithm which we call SVM with Gaussian Sample Uncertainty (SVM-GSU). In Section~\ref{sec:experiments}, we provide the experimental results of the application of SVM-GSU to synthetic data and to five publicly available and popular datasets. In the same section, we provide comparisons with the standard SVM and other state of the art methods. In Section~\ref{sec:conclusion}, we draw some conclusions and give directions for future work.

\section{Related Work}\label{sec:rel_work}

	Uncertainty is ubiquitous in almost all fields of scientific studies~\cite{li2013dealing}. Exploiting uncertainty in supervised learning has been studied in many different aspects~\cite{deisenroth2015gaussian, bengio2013representation, joshi2009multi}. More specifically, the research community has studied learning problems where uncertainty is present either in the labels or in the representation of the training data.
	
	In~\cite{liu2016classification}, Liu and Tao studied a classification problem in which sample labels are randomly corrupted. In this scenario, there is an unobservable sample with noise-free labels. However, before being observed, the true labels are independently flipped with a probability $p\in[0,0.5)$, and the random label noise can be class-conditional. Tzelepis et al.~\cite{tzelepis2013improving, tzelepis2016learning} proposed an SVM extension where each training example is assigned a relevance degree in $(0,1]$ expressing the confidence that the respective example belongs to the given class. Li and Sethi~\cite{li2006confidence} proposed an active learning approach based on identifying and annotating uncertain samples. Their approach estimates the uncertainty value for each input sample according to its output score from a classifier and selects only samples with uncertainty value above a user-defined threshold. In~\cite{sarafis2015building}, the authors used weights to quantify the confidence of automatic training label assignment to images from clicks and showed that using these weights with Fuzzy SVM and Power SVM~\cite{zhang2012power} can lead to significant improvements in retrieval effectiveness compared to the standard SVM. Finally, the problem of confidence-weighted learning is addressed in~\cite{crammer2009adaptive, dredze2008confidence, hoi2008exact}, where uncertainty in the weights of a linear classifier (under online learning conditions) is taken into consideration.
	    
	Assuming uncertainty in data representation has also drawn the attention of the research community in recent years. Different types of robust SVMs have been proposed in several recent works. Bi and Zhang~\cite{bi2004support} considered a statistical formulation where the input noise is modeled as a hidden mixture component, but in this way the ``iid'' assumption for the training data is violated. In that work, the uncertainty is modeled isotropically. Second order cone programming (SOCP)~\cite{alizadeh2003second} methods have also been employed in numerous works to handle missing and uncertain data. In addition, Robust Optimization techniques~\cite{ben1998robust,bertsimas2011theory} have been proposed for optimization problems where the data is not specified exactly, but it is known to belong to a given uncertainty set $\mathcal{U}$, yet the optimization constraints must hold for all possible values of the data from $\mathcal{U}$.
	    
	Lanckriet et al.~\cite{lanckriet2003robust} considered a binary classification problem where the mean and covariance matrix of each class are assumed to be known. Then, a minimax problem is formulated such that the worst-case (maximum) probability of misclassification of future data points is minimized. That is, under all possible choices of class-conditional densities with a given mean and covariance matrix, the worst-case probability of misclassification of new data is minimized.
	    
	Shivaswamy et al.~\cite{shivaswamy2006second}, who extended Bhattacharyya et al.~\cite{bhattacharyya2004second}, also adopted a SOCP formulation and used generalized Chebyshev inequalities to design robust classifiers dealing with uncertain observations. In their work uncertainty arises in ellipsoidal form, as follows from the multivariate Chebyshev inequality. This formulation achieves robustness by requiring that the ellipsoid of every uncertain data point should lie in the correct halfspace. The expected error of misclassifying a sample is obtained by computing the volume of the ellipsoid that lies on the wrong side of the hyperplane. However, this quantity is not computed analytically; instead, a large number of uniformly distributed points are generated in the ellipsoid, and the ratio of the number of points on the wrong side of the hyperplane to the total number of generated points is computed.
	    
	Several works~\cite{bhattacharyya2004second,shivaswamy2006second,lanckriet2003robust} robustified regularized classification using box-type uncertainty. By contrast, Xu et al.~\cite{xu2009robustness, xu2012robustness} considered the robust classification problem for a class of non-box-typed uncertainty sets; that is, they considered a setup where the joint uncertainty is the Cartesian product of uncertainty in each input. This leads to penalty terms on each constraint of the resulting formulation. Furthermore, Xu et al. gave evidence on the equivalence between the standard regularized SVM and this robust optimization formulation, establishing robustness as the \textit{reason} why regularized SVMs generalize well.
	    
	In~\cite{qi2013robust}, motivated by GEPSVM~\cite{mangasarian2006multisurface}, Qi et al. robustified a twin support vector machine (TWSVM)~\cite{khemchandani2007twin}. Robust TWSVM~\cite{qi2013robust} deals with data affected by measurement noise using a SOCP formulation. In their work, the input data is contaminated with isotropic noise (i.e., spherical disturbances centred at the training examples), and thus cannot model real-world uncertainty, which is typically described by more complex noise patterns. Power SVM~\cite{zhang2012power} uses a spherical uncertainty measure for each training example. In this formulation, each example is represented by a spherical region in the feature space, rather than a point. If any point of this region is classified correctly, then the corresponding loss introduced is zero.

	Our proposed classifier does not violate the ``iid'' assumption for the training input data (in contrast to~\cite{bi2004support}), and can model the uncertainty of each input training example using an arbitrary covariance matrix; that is, it allows anisotropic modeling of the uncertainty analytically in contrast to~\cite{shivaswamy2006second, qi2013robust, zhang2012power}. Moreover, we define a cost function that is convex and whose derivatives with respect to the parameters of the unknown separating hyperplane can be expressed in closed form. Therefore, we can find their global optimal using an iterative gradient descent algorithm whose complexity is linear with respect to the number of training data. Finally, we apply a linear subspace learning approach in order to address the situation where most of the mass of the Gaussians lies in a low dimensional manifold that can be different for each Gaussian, and subsequently solve the problem in lower-dimensional spaces. Learning in subspaces is widely used in various statistical learning problems~\cite{de2003framework, liwicki2012efficient, lu2009uncorrelated}.

\section{Proposed Approach}\label{sec:prop_approach}
	
	In this section we develop a new classification algorithm whose training set is not just a set of vectors $\mathbf{x}_i$ in some multi-dimensional space, but rather a set of multivariate Gaussian distributions; that is, each training example consists of a mean vector $\mathbf{x}_i\in\mathcal{D}$ and a covariance matrix $\Sigma_i\in\mathbb{S}_{++}^{n}$; the latter expresses the uncertainty around the corresponding mean\footnote{$\mathcal{D}$ is typically a subset of the $n$-dimensional Euclidean space of column vectors, while $\mathbb{S}_{++}^{n}$ denotes the convex cone of all symmetric positive definite $n\times n$ matrices with entries in $\mathcal{D}\subseteq\mathbb{R}^n$.}. In Sect.~\ref{ssec:svmgsu}, we first briefly review the linear SVM and then describe in detail the proposed linear SVM with Gaussian Sample Uncertainty (SVM-GSU). In Sect.~\ref{ssec:svmgsu_sl} we motivate and describe a formulation that allows learning in linear subspaces. In the general case we arrive at different subspaces for the different Gaussians -- this allows, for example, dealing with covariance matrices that are of low rank. In Sect.~\ref{ssec:sampling} we discuss how the proposed algorithm relates to standard SVM when the latter is fed with samples drawn from the input Gaussians. Finally, in Sect.~\ref{ssec:sgd} we describe a SGD algorithm for efficiently solving the SVM-GSU optimization problem.
	
	\subsection{SVM with Gaussian Sample Uncertainty}\label{ssec:svmgsu}
	
\begin{figure*}[t]
			\centering
			\begin{subfigure}[b]{0.30\linewidth}
			      \includegraphics[width=\linewidth]{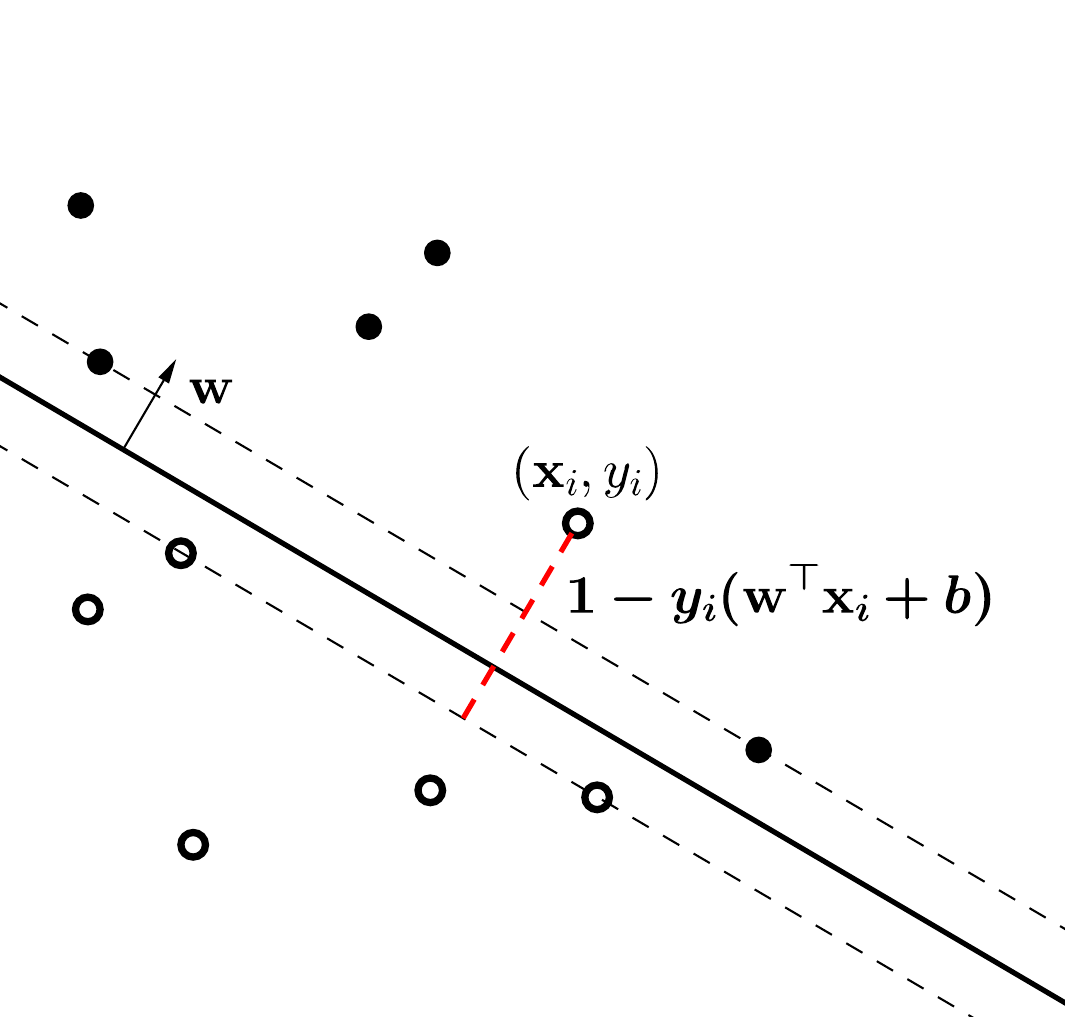}
			      \caption{}
			      \label{subfig:standard_svm}
			\end{subfigure}
			~
			\begin{subfigure}[b]{0.30\linewidth}
			      \includegraphics[width=\linewidth]{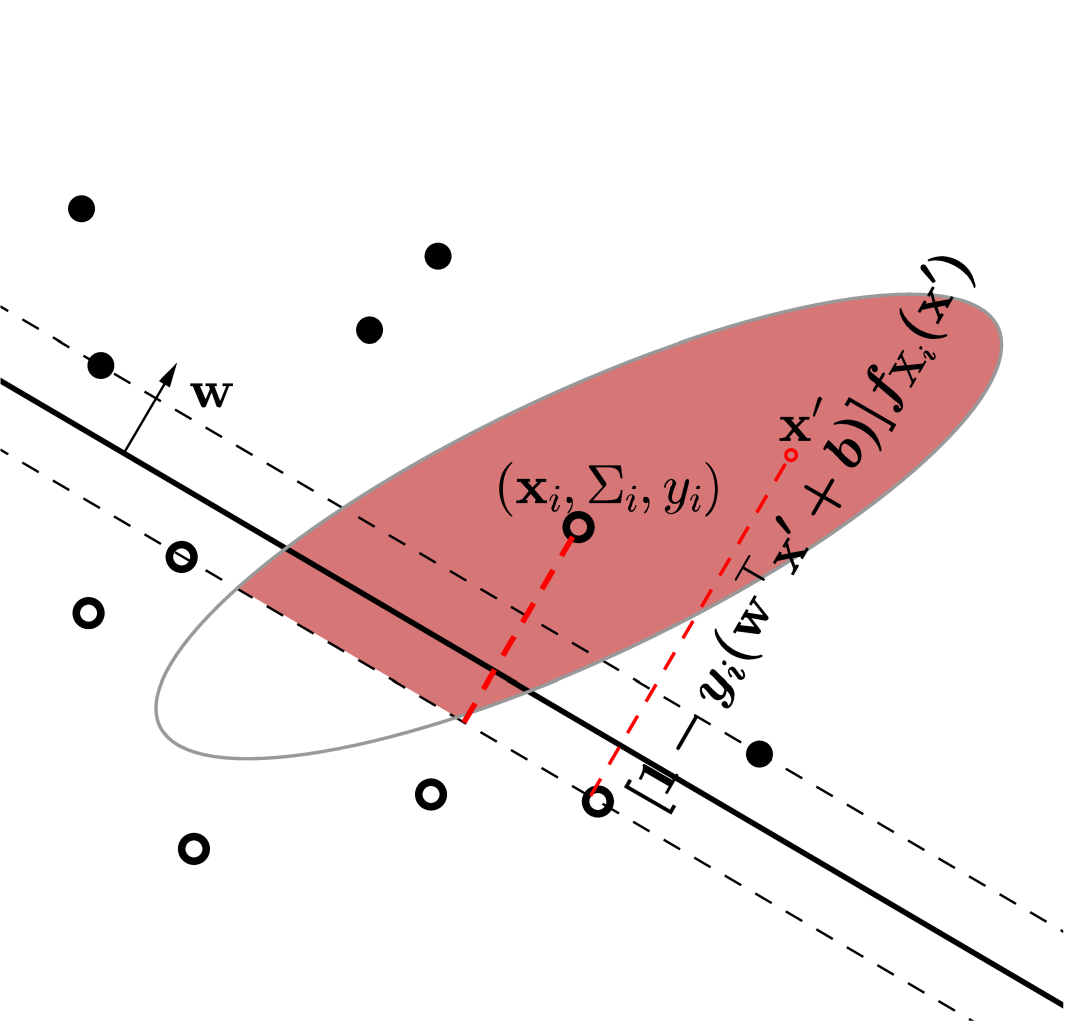}
			      \caption{}
			      \label{subfig:svmgsu}
			\end{subfigure}
			~
			\begin{subfigure}[b]{0.30\linewidth}
			      \includegraphics[width=\linewidth]{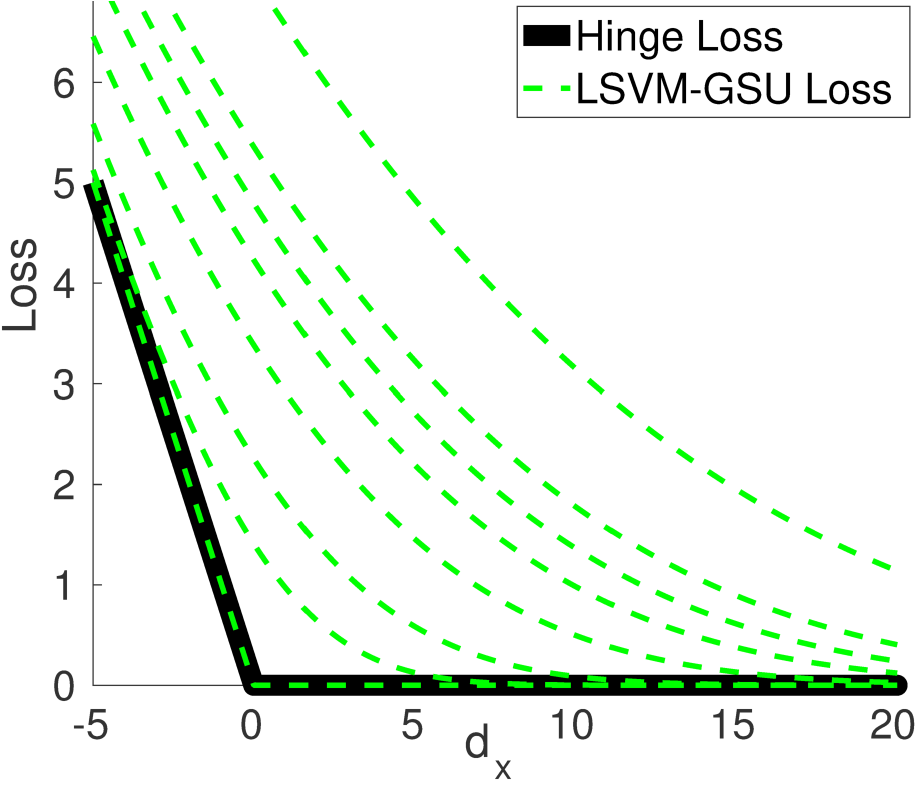}
			      \caption{}
			      \label{subfig:losses}
			\end{subfigure}
			\caption{Illustrative example of calculating (a) the standard linear SVM's hinge loss, and (b) the proposed linear SVM-GSU's loss. In (c), the hinge loss is compared with the proposed linear SVM-GSU's loss for various quantities of uncertainty.}
			\label{fig:svm_vs_svmgsu}
		\end{figure*}

		We begin by briefly describing the standard SVM algorithm. Let us consider the supervised learning framework and denote the training set with $\mathcal{X}=\big\{(\mathbf{x}_i, y_i) \colon \mathbf{x}_i\in\mathbb{R}^{n},\:y_i\in\{\pm1\}, i=1,\ldots,\ell\big\}$, where $\mathbf{x}_i$ is a training example and $y_i$ is the corresponding class label. Then, the standard linear SVM learns a hyperplane $\mathcal{H}\colon\mathbf{w}^\top\mathbf{x}+b=0$ that minimizes with respect to $\mathbf{w}$, $b$ the following objective function:
		\begin{equation}\label{eq:svm_obj}
			\frac{\lambda}{2}\lVert\mathbf{w}\rVert^2 + \frac{1}{\ell}\sum_{i=1}^{\ell}\max\left(0,1-y_i(\mathbf{w}^\top\mathbf{x}_i+b)\right),
		\end{equation}
		where $h(t)=\max(0,1-t)$ is the ``hinge'' loss function~\cite{hastie2004entire}. An illustrative example of the hinge loss calculation is given in Fig.~\ref{fig:svm_vs_svmgsu}, where in Fig.~\ref{subfig:standard_svm} the red dashed line indicates the loss introduced by the misclassified example $(\mathbf{x}_i,y_i)$ and in Fig.~\ref{subfig:losses} the hinge loss is shown in the black bold line.
		  
		In this work we assume that, instead of the $i$-th training example in the form of a vector, we are given a multivariate Gaussian distribution with mean vector $\mathbf{x}_i$ and covariance matrix $\Sigma_i$. One could think of this as that the covariance matrix, $\Sigma_i$, models the uncertainty about the position of training samples around $\mathbf{x}_i$. Formally, our training set is a set of $\ell$ annotated Gaussian distributions, i.e., $\mathcal{X}'=\big\{ (\mathbf{x}_i, \Sigma_i, y_i) \colon \mathbf{x}_i\in\mathbb{R}^{n}, \Sigma_i\in\mathbb{S}_{++}^n, y_i\in\{\pm1\},\:i=1,\ldots,\ell\big\}$, where $\mathbf{x}_i\in\mathbb{R}^n$ and  $\Sigma_i\in\mathbb{S}_{++}^n$ are respectively the mean vector and the covariance matrix of the $i$-th example, and $y_i$ is the corresponding label. Then, we define $\ell$ random variables, $\mathbf{X}_i$, each of which we assume that follows the corresponding $n$-dimensional Gaussian distribution $\mathcal{N}(\mathbf{x}_i,\Sigma_i)$ and define an optimization problem where the misclassification cost for the $i$-th example is the expected value of the hinge loss for the corresponding Gaussian. Formally, the optimization problem, in its unconstrained primal form, is the minimization with respect to $\mathbf{w}$, $b$ of
		\begin{eqnarray}\label{eq:svmgsu_obj_int}
			\frac{\lambda}{2}\lVert\mathbf{w}\rVert^2 + 
			\frac{1}{\ell}\sum_{i=1}^{\ell}
			\bigintssss_{\mathbb{R}^n} 
			\!
			\max\big(0, 1-y_i(\mathbf{w}^\top\mathbf{x}+b)\big)
			f_{\mathbf{X}_i}(\mathbf{x})
			\, \mathrm{d}\mathbf{x},\mkern-10mu
		\end{eqnarray}
		where $f_{\mathbf{X}_i}(\mathbf{x}) = \frac{1}{(2\pi)^{\frac{n}{2}}|\Sigma_i|^{\frac{1}{2}}}\exp\left(-\frac{1}{2}(\mathbf{x}-\mathbf{x}_i)^\top\Sigma^{-1}_i(\mathbf{x}-\mathbf{x}_i)\right)$ is the probability density function (PDF) of the $i$-th Gaussian distribution. The above objective function $\mathcal{J}\colon\mathbb{R}^n\times\mathbb{R}\to\mathbb{R}$ can be written as
		\begin{equation}\label{eq:svmgsu_obj_loss}
			\mathcal{J}(\mathbf{w},b)=
			\frac{\lambda}{2}\lVert\mathbf{w}\rVert^2
			+
			\frac{1}{\ell}\sum_{i=1}^{\ell}
			\mathcal{L}\big(\mathbf{w},b;(\mathbf{x}_i,\Sigma_i,y_i)\big),
		\end{equation}
		where, as stated above, the loss function $\mathcal{L}$ for the $i$-th example (i.e. the $i$-th Gaussian) is defined as the expected value of the hinge loss for the Gaussian in question. That is,
		\begin{equation}\label{eq:svmgsu_loss_int}
			\mathcal{L}(\mathbf{w},b)
			\mkern-5mu = \mkern-10mu
			\bigintssss_{\mathbb{R}^n} 
			\mkern-15mu
			\max\big(0, 1\mkern-6mu-y_i(\mathbf{w}^\top\mathbf{x}+b)\big)
			f_{\mathbf{X}_i}(\mathbf{x})
			\, \mathrm{d}\mathbf{x}.
		\end{equation}
		  
		We proceed to express the objective function (\ref{eq:svmgsu_obj_loss}) and its derivatives in closed form. This will allow us to solve the corresponding optimization problem using an efficient SGD approach. More specifically, the loss can be expressed as
		\begin{equation}\label{eq:svmgsu_loss_int_omega}
			\mathcal{L}(\mathbf{w},b)
			\mkern-5mu = \mkern-5mu
			\bigintssss_{\Omega_i} 
			\!
			\left[1-y_i(\mathbf{w}^\top\mathbf{x}+b)\right]
			f_{\mathbf{X}_i}(\mathbf{x})
			\, 
			\mathrm{d}\mathbf{x},
		\end{equation}
		where $\Omega_i$ denotes the halfspace of $\mathbb{R}^n$ that is defined by the hyperplane $\mathcal{H}'\colon y_i(\mathbf{w}^\top\mathbf{x}+b)=1$ as $\Omega_i=\{\mathbf{x}\in\mathbb{R}^n\colon y_i(\mathbf{w}^\top\mathbf{x}+b)\leq 1\}$, and is the halfspace to which misclassified samples lie. This is illustrated in Fig.~\ref{subfig:svmgsu}, where a misclassified example $(\mathbf{x}_i,\Sigma_i,y_i)$ introduces a loss indicated by the shaded region. For the calculation of this loss, all points that belong to the halfspace $\Omega_i=\{\mathbf{x}\in\mathbb{R}^n\colon y_i(\mathbf{w}^\top\mathbf{x}+b)\leq 1\}$, i.e., the points $\mathbf{x}^\prime\in\Omega_i$, contribute to it by a quantity of $[1-y_i(\mathbf{w}^\top\mathbf{x}^\prime+b)]f_{\mathbf{X}_i}(\mathbf{x}^\prime)$. For one such $\mathbf{x}^\prime$ denoted by a red circle in Fig.~\ref{subfig:svmgsu}, the first part of the above product, $1-y_i(\mathbf{w}^\top\mathbf{x}^\prime+b)$, corresponds to the typical hinge loss of SVM, shown as a red dashed line in this example. The total loss introduced by the misclassified example $(\mathbf{x}_i,\Sigma_i,y_i)$ is obtained by integrating all these quantities over the halfspace $\Omega_i$. 
		
		Using Theorem~\ref{thm:thm_1} proved in Appendix~\ref{app:app_A}, for the halfspace $\Omega_i^+ = \left\{\mathbf{x}\in\mathbb{R}^n \colon y_i\left(\mathbf{w}^\top\mathbf{x}+b\right)\leq1\right\}$, the above integral is evaluated in terms of $\mathbf{w}$ and $b$ as follows
		\begin{equation}\label{eq:svmgsu_loss}
			\mathcal{L}(\mathbf{w},b) =  
			\frac{d_{\mathbf{x}_i}}{2}
			\left[\operatorname{erf}\left(\frac{d_{\mathbf{x}_i}}{d_{\Sigma_i}}\right)+1\right]
			+
			\frac{d_{\Sigma_i}}{2\sqrt{\pi}}
			\exp\left(-\frac{d_{\mathbf{x}_i}^2}{d_{\Sigma_i}^2}\right),
		\end{equation}
		where $d_{\mathbf{x}_i}=1-y_i\left(\mathbf{w}^\top\mathbf{x}_i+b\right)$, $d_{\Sigma_i}=\sqrt{2\mathbf{w}^\top\Sigma_i\mathbf{w}}$, and $\operatorname{erf}\colon\mathbb{R}\to(-1,1)$ is the error function, defined as $\operatorname{erf}(x)=\frac{2}{\sqrt{\pi}}\int_{0}^{x}\!e^{-t^2}\,\mathrm{d}t$. For a training example $(\mathbf{x},\Sigma,y)$, Fig.~\ref{subfig:losses} shows the proposed loss in dashed green lines for constant values of $d_\Sigma$ (constant amounts of uncertainty). We note that as $d_\Sigma\rightarrow0$, SVM-GSU's loss virtually coincides with the SVM's hinge loss, while it can be easily verified that, regardless of $d_\Sigma$, as $d_\mathbf{x}\rightarrow\infty$ the SVM-GSU's loss will eventually converge to zero (as the hinge loss does).
		  
		Let us note that the covariance matrix of each training example describes the uncertainty around the corresponding mean; that is, as the covariance matrix approaches the zero matrix, the certainty increases. At the extreme\footnote{A zero covariance matrix exists due to the well known property that the set of symmetric positive definite matrices is a convex cone with vertex at zero.}, as $\Sigma\to\mathbf{0}$, the proposed loss converges to the hinge loss function used in the standard SVM formulation~\cite{hastie2004entire}. This implies that the proposed formulation is a generalization of the standard SVM; the two classifiers are equivalent when the covariance matrices tend to the zero matrix.
		
		It is easy to show that the objective function (\ref{eq:svmgsu_obj_loss}) is convex with respect to $\mathbf{w}$ and $b$; therefore, we propose a SGD algorithm in Sect.~\ref{ssec:sgd} for solving the corresponding optimization problem. Since the objective function is convex, we can obtain the global optimal solution. Moreover, it can be shown that the proposed loss function (\ref{eq:svmgsu_loss_int}) enjoys the consistency property~\cite{rosasco2004loss, zhang2004statistical}, i.e., it leads to consistent results with the $0-1$ loss given the presence of infinite data. By differentiating $\mathcal{J}$ with respect to $\mathbf{w}$ and $b$, we obtain, respectively,
		\begin{eqnarray}\label{eq:dJ_dw}
			\frac{\partial\mathcal{J}}{\partial\mathbf{w}}
			= 
			\lambda\mathbf{w} +     
			\frac{1}{\ell}\sum_{i=1}^{\ell}
			\Bigg[
			\frac{\exp\left(-\sfrac{d_{\mathbf{x}_i}^2}{d_{\Sigma_i}^2}\right)}{\sqrt{\pi}d_{\Sigma_i}}
			\Sigma_i\mathbf{w} 
			\quad\quad\quad\quad\quad
			\nonumber \\
			- 
			\frac{1}{2}\left(
			\operatorname{erf}\left(\frac{d_{\mathbf{x}_i}}{d_{\Sigma_i}}\right)+1\right)\mathbf{x}_i
			\Bigg],
		\end{eqnarray}
		\begin{equation}\label{eq:dJ_db}
			\frac{\partial\mathcal{J}}{\partial b}
			= 
			-\frac{1}{\ell}\sum_{i=1}^{\ell}
			\left[\operatorname{erf}\left(\frac{d_{\mathbf{x}_i}}{d_{\Sigma_i}}\right)+1\right].
		\end{equation}

		Despite the complex appearance of the loss function and its derivatives, their computation essentially requires the calculation of the inner product $\mathbf{w}^\top\mathbf{x}_i$ (which is the same as in standard SVM), plus that of the quadratic form $\mathbf{w}^\top\Sigma_i\mathbf{w}$, which requires $\frac{n(n+1)}{2}$ multiplications, since $\Sigma_i$ is symmetric. The latter, in the case of diagonal covariance matrices, is equivalent to the computation of an inner product, i.e., of complexity $\mathcal{O}(n)$. Moreover, each one of $\mathbf{w}^\top\mathbf{x}_i$ and $\mathbf{w}^\top\Sigma_i\mathbf{w}$ needs to be computed just once for calculating the loss function and its derivatives for a given $\mathbf{w}$. It is worth noting that, in practice, as shown in Sect.~\ref{sec:experiments}, in real-world problems uncertainty usually rises in diagonal form. In such cases, the proposed algorithm is quite efficient and exhibits very similar complexity to the standard linear SVM.

		Once the optimal values of the parameters $\mathbf{w}$ and $b$ are learned, an unseen testing datum, $\mathbf{x}_t$, can be classified to one of the two classes according to the sign of the (signed) distance between $\mathbf{x}_t$ and the separating hyperplane. That is, the predicted label of $\mathbf{x}_t$ is computed as $y_t=\operatorname{sgn}(d_t)$, where $d_t=(\mathbf{w}^\top\mathbf{x}_t+b)/\lVert\mathbf{w}\rVert$. The posterior class probability, i.e, a probabilistic degree of confidence that the testing sample belongs to the class to which it has been classified, can be calculated using the well-known Platt scaling approach~\cite{platt1999probabilistic} for fitting a sigmoid function, $S(t)=1/(1+e^{\sigma_At+\sigma_B})$. This is the same approach that is used in the standard linear SVM formulation (e.g., see~\cite{CC01a}) for evaluating a sample's class membership at the testing phase.

	\subsection{Solving the SVM-GSU in linear subspaces}\label{ssec:svmgsu_sl}

		The derivations in Sect.~\ref{ssec:svmgsu} were made for the general case of full rank covariance matrices that can be different for each of the examples. Clearly, one can introduce constraints on the covariance matrices, such as them being diagonal, block diagonal, or multiples of the identity matrix. In this way one can model different types of uncertainty -- examples will be given in the section of experimental results. However, in some cases, especially when the dimensionality of the data is high, most of the mass of the Gaussian distributions will lie in a few directions in the feature space that may be different for each example and may not be aligned with the feature axes. To address this issue we alter the formulation and work directly in the subspaces that preserve most of the variance. More specifically, we propose a methodology for approximating the loss function of SVM-GSU, by projecting the vectors $\mathbf{x}$ in (\ref{eq:svmgsu_loss_int_omega}) into a linear subspace and integrating the hinge loss function in that subspace instead of the original feature space. A separate subspace is used for each of the training examples, that is, for each of the input Gaussians. For a given Gaussian distribution, the projection matrix is found by performing eigenanalysis on the covariance matrix and the dimensionality of each subspace is defined so as to preserve a certain fraction of the total variance.

		More specifically, by performing eigenanalysis on the covariance matrix of the random vector $\mathbf{X}_i$, the latter is decomposed as $\Sigma_i = U_i \Lambda_i U_i^\top$, where $\Lambda_i$ is an $n\times n$ diagonal matrix consisting of the eigenvalues of $\Sigma_i$, i.e. $\Lambda_i=\operatorname{diag}(\lambda_i^1,\ldots,\lambda_i^n)$, so that $\lambda_i^1\geq,\ldots,\geq\lambda_i^n>0$, while $U_i$ is an $n\times n$ orthonormal matrix, whose $j$-th column, $\mathbf{u}_i^j$, is the eigenvector corresponding to the $j$-th eigenvalue, $\lambda_i^j$.
		  
		Let us keep the first $d_i\leq n$ eigenvectors, so that a certain fraction $p\in(0,1]$ of the total variance is preserved, i.e., $\frac{\sum_{t=1}^{d_i}\lambda_i^t}{\sum_{t=1}^{n}\lambda_i^t}>p$. Then, we construct the $n\times d_i$ matrix $U_i'$ by keeping the first $d_i$ columns of $U_i$, i.e., $U_i^\prime = [\mathbf{u}_i^1\:\:\mathbf{u}_i^2\:\:\ldots\:\:\mathbf{u}_i^{d_i}]$. Now, by using the projection matrix $P_i={U_i^\prime}^\top$, we define a new random vector $\mathbf{Z}_i$, such that $\mathbf{Z}_i = P_i\mathbf{X}_i$. Then, $\mathbf{Z}_i\in\mathbb{R}^{d_i}$ follows a multivariate Gaussian distribution (since $\mathbf{X}_i\sim\mathcal{N}(\mathbf{x}_i,\Sigma_i)$), i.e. $\mathbf{Z}_i\sim\mathcal{N}(\mathbf{z}_i,\Sigma_i^z)$, with mean vector $\mathbf{z}_i = \mathbb{E}\big[P_i\mathbf{X}_i\big] = P_i\mathbb{E}\big[\mathbf{X}_i\big] = P_i\mathbf{x}_i$ and (diagonal) covariance matrix $\Sigma_i^z = \Lambda_i^z$. Let $f_{\mathbf{Z}_i}$ denote the PDF of $\mathbf{Z}_i$.
		  
		We proceed to approximate the expected value of the hinge loss in the original space (\ref{eq:svmgsu_loss_int_omega}), by considering the integral in the new, lower-dimensional space where most of the variance is preserved. More specifically, $\mathbf{x} \approx P_i^\top\mathbf{z}\implies\mathbf{w}^\top\mathbf{x}\approx\mathbf{w}^\top(P_i^\top\mathbf{z})=\mathbf{w_z}^\top\mathbf{z}$, where $\mathbf{w}_z = P_i\mathbf{w}$. Consequently, the loss function for the $i$-th example, that is the integral in the RHS of (\ref{eq:svmgsu_loss_int_omega}) can be approximated by the quantity $\bigintssss_{\Omega_i^z}\!\left[1-y_i\big(\mathbf{w}_z^\top\mathbf{z}+b\big)\right]f_{\mathbf{Z}_i}(\mathbf{z})\,\mathrm{d}\mathbf{z},$ where $\Omega_i^z$ denotes the projected halfspace on $\mathbb{R}^{d_i}$, that is, $\Omega_i^z=\big\{\mathbf{z}\in\mathbb{R}^{d_i}\colon y_i\big(\mathbf{w}_z^\top\mathbf{z}+b\big)\leq1\big\}$. Using Theorem~\ref{thm:thm_1} (Appendix~\ref{app:app_A}), we can then give this approximation of the loss function $\mathcal{L}^\prime\colon\mathbb{R}^{d_i}\times\mathbb{R}\to\mathbb{R}$, in closed form as follows:
		\begin{equation}\label{eq:svmgsu_loss_z}
			\mathcal{L}^\prime(\mathbf{w},b) =  
			\frac{d_{\mathbf{z}_i}}{2}
			\left[\operatorname{erf}\left(\frac{d_{\mathbf{z}_i}}{d_{\Sigma_i^z}}\right)+1\right]
			+
			\frac{d_{\Sigma_i^z}}{2\sqrt{\pi}}
			\exp\left(-\frac{d_{\mathbf{z}_i}^2}{d_{\Sigma_i^z}^2}\right)
		\end{equation}
		where $d_{\mathbf{z}_i}=1-y_i\left(\mathbf{w}_z^\top\mathbf{z}_i+b\right)$, $d_{\Sigma_i^z}=\sqrt{2\mathbf{w}_z^\top\Sigma_i^z\mathbf{w}_z}$. Therefore, the objective function $\mathcal{J}^\prime\colon\mathbb{R}^n\times\mathbb{R}\to\mathbb{R}$, given by (\ref{eq:svmgsu_obj_loss}) can be approximated as follows
		\begin{equation}\label{eq:svmgsu_obj_loss_z}
			\mathcal{J}^\prime(\mathbf{w},b)
			=
			\frac{\lambda}{2}\|\mathbf{w}\|^2
			+
			\frac{1}{\ell}\sum_{i=1}^{\ell}
			\mathcal{L}^\prime\left(P_i\mathbf{w},b;(\mathbf{z}_i,\Sigma_i^z,y_i)\right).
		\end{equation}
		Similarly to $\mathcal{J}$, we can show that $\mathcal{J}^\prime$ is also convex with respect to the unknown parameters $\mathbf{w}$ and $b$ of the separating hyperplane. Moreover, using the chain rule, we can obtain the partial derivatives of $\mathcal{J}^\prime$ with respect to $\mathbf{w}$ and $b$ in closed form, and therefore use a stochastic gradient method to arrive at the global optimum. More specifically,
		$$
			\frac{\partial\mathcal{J'}}{\partial\mathbf{w}} =
			\lambda\mathbf{w} +     
			\frac{1}{\ell}\sum_{i=1}^{\ell}
			\frac{\partial}{\partial\mathbf{w}_z}
			\mathcal{L}^\prime\big(\mathbf{w}_z,b;(\mathbf{z}_i,\Sigma_i^z,y_i)\big)
			\frac{\partial\mathbf{w}_z}{\partial\mathbf{w}},
		$$
		where $\frac{\partial}{\partial\mathbf{w}}\mathbf{w}_z=\frac{\partial}{\partial\mathbf{w}}P_i\mathbf{w} = P_i$. By differentiating $\mathcal{L}^\prime$ with respect to $\mathbf{w}_z$, and replacing in the above, we arrive at
		\begin{eqnarray}\label{eq:dJprime_dw}
			\frac{\partial\mathcal{J}^\prime}{\partial\mathbf{w}}
			= 
			\lambda\mathbf{w} +     
			\frac{1}{\ell}\sum_{i=1}^{\ell}
			\Bigg[
			\frac{\exp\left(-\sfrac{d_{\mathbf{z}_i}^2}{d_{\Sigma_i^z}^2}\right)}{\sqrt{\pi}d_{\Sigma_i^z}}
			P_i^\top(\Sigma_i^z\mathbf{w}_z)
			\quad\quad
			\nonumber \\
			- 
			\frac{1}{2}\left(
			\operatorname{erf}\left(\frac{d_{\mathbf{z}_i}}{d_{\Sigma_i^z}}\right)+1\right)P_i^\top\mathbf{z}_i
			\Bigg],
		\end{eqnarray}
		that is a closed form equation that gives the partial derivatives of the cost with respect to $\mathbf{w}$. Similarly, the first partial derivative of $\mathcal{J}^\prime$ with respect to $b$ can be obtained as follows
		\begin{equation}\label{eq:dJprime_db}
			\frac{\partial\mathcal{J'}}{\partial b}
			= 
			-\frac{1}{\ell}\sum_{i=1}^{\ell}
			\left[\operatorname{erf}\left(\frac{d_{\mathbf{z}_i}}{d_{\Sigma_i^z}}\right)+1\right].
		\end{equation}
		where $\mathbf{w}_z = P_i\mathbf{w}$, $\Sigma_i^z = P_i \Sigma_i P_i^\top$.

		To summarize, in the low-dimensional spaces $\mathbb{R}^{d_i}$, the loss function is computed as shown in (\ref{eq:svmgsu_loss_z}). The objective function is computed as shown in (\ref{eq:svmgsu_obj_loss_z}) and its first derivatives are computed as in (\ref{eq:dJprime_dw}) and (\ref{eq:dJprime_db}). Finally, let us note that in the above equations, the only matrix operations involve the projection matrix $P_i$. Since the covariance matrices $\Sigma_i^z$ are diagonal, all operations that involve them boil down to efficient vector rescaling and vector norm calculations.

	\subsection{To sample or not to sample?}\label{ssec:sampling}
	    
		The data term in our formulation (see (\ref{eq:svmgsu_loss_int})) is the expected value of the classical SVM cost when data samples are drawn from the multi-dimensional Gaussian distributions. It therefore follows that a standard linear SVM would arrive at the same hyperplane when sufficiently many samples are drawn from them. How many samples are needed to arrive at the same hyperplane is something that cannot be computed analytically. Nevertheless, our analysis and results indicate that this number can be prohibitively high, especially in the case of high-dimensional spaces. 

		More specifically, in what follows, we show that the difference between the analytically calculated expected value of the hinge loss (\ref{eq:svmgsu_loss_int}) and its sample mean is bounded by a quantity that is inversely related to the dimensionality of the feature space. Let $\mathcal{L}$ be the expected loss given analytically as in (\ref{eq:svmgsu_loss_int}), and $\tilde{\mathcal{L}}_N$ its approximation when $N$ samples are drawn from the Gaussians. Since the hinge loss is $\rVert\mathbf{w}\rVert$-Lipschitz\footnote{A function $h\colon\mathbb{R}^n\to\mathbb{R}$ is $\mathscr{L}$-Lipschitz with respect to the Euclidean norm if $\lvert h(\mathbf{x})-h(\mathbf{y})\rvert\leq\mathscr{L}\lVert\mathbf{x}-\mathbf{y}\rVert$, $\mathscr{L}>0$. Indeed, the hinge loss $h(\mathbf{x})=\max(0,1-y(\mathbf{w}^\top\mathbf{x}+b))$ is $\rVert\mathbf{w}\rVert$-Lipschitz since $\left\lvert h(\mathbf{x})-h(\mathbf{y})\right\rvert \leq \left\lvert 1-y(\mathbf{w}^\top\mathbf{x}+b)-1+y(\mathbf{w}^\top\mathbf{y}+b) \right\rvert \leq \rVert\mathbf{w}\rVert \lVert\mathbf{x}-\mathbf{y}\lVert$.} with respect to the Euclidean norm, we can use a result due to Tsirelson et al.~\cite{Tsirelson1976} that provides a concentration inequality for Lipschitz functions of Gaussian variables. By doing so, for all $r\geq0$, we arrive at the following concentration inequality
		\begin{equation}
			P\left(\left\lvert \mathcal{L} - \tilde{\mathcal{L}}_N \right\rvert \geq r\right)
			\leq
			2\exp\left(-\frac{r^2}{2\rVert\mathbf{w}\rVert^2}\right).
		\end{equation}
		That is, the tails of the error probability decay exponentially with $r^2$. More interestingly, they increase with the squared norm of $\lVert\mathbf{w}\rVert$, and therefore with the dimensionality of the input space, $n$. Consequently, as $n$ increases, one needs to generate more samples from the Gaussians in order to preserve a desired approximation of the loss.
		  
		This means that for spaces of high dimensionality the number of samples needed to approximate (\ref{eq:svmgsu_loss_int}) sufficiently well, can be prohibitively high. We experimentally demonstrated this with a toy example in Sect.~\ref{ssec:exp_toy} (see Fig.~\ref{fig:thetas}), where we show that in $2$ dimensions we need approximately $3$ orders of magnitude more samples to arrive at the same hyperplane, while for $3$ dimensions we need $4$ orders of magnitude more samples. Our experimental results on the large-scale MED dataset (Sect.~\ref{ssec:exp_med}) also show the limitations of a sampling approach.

	\subsection{A stochastic gradient descent solver for SVM-GSU}\label{ssec:sgd}
		
		Motivated by the Pegasos algorithm (Primal Estimated sub-GrAdient SOlver for SVM), first proposed by Shalev-Shwartz et al. in~\cite{shalev2011pegasos}, we present a stochastic sub-gradient descent algorithm for solving SVM-GSU in order to efficiently address scalability requirements\footnote{A C++ implementation of the proposed method can be found at \url{https://github.com/chi0tzp/svm-gsu}.}.
		  
		Pegasos is a well-studied algorithm~\cite{shalev2011pegasos, kakade2009generalization} providing both state of the art classification performance and great scalability. It requires $\tilde{\mathcal{O}}(\sfrac{1}{\epsilon})$ number of iterations in order to obtain a solution of accuracy $\epsilon$, in contrast to previous analyses of SGD methods that require $\tilde{\mathcal{O}}(d/(\lambda\epsilon))$ iterations, where $d$ is a bound on the number of non-zero features in each example\footnote{We use the $\tilde{\mathcal{O}}$ notation (soft-O) as a shorthand for the variant of $\mathcal{O}$ (big-O) that ignores logarithmic factors; that is, $f(n)\in\tilde{\mathcal{O}}(g(n))\iff\exists k\in\mathbb{N}\colon f(n)\in\mathcal{O}(g(n)\log^k(g(n)))$.}. Since the run-time does not depend directly on the size of the training set, the resulting algorithm is especially suited for learning from large datasets.
		  
		Given a training set $\mathcal{X}=\big\{(\mathbf{x}_i,\Sigma_i,y_i)\colon\mathbf{x}_i\in\mathbb{R}^{n},\Sigma_i\in\mathbb{S}_{++}^n,y_i\in\{\pm1\},\:i=1,\ldots,\ell\big\}$, the proposed algorithm solves the following optimization problem
		\begin{equation}
			\displaystyle \min_{\mathbf{w},b}
			\frac{\lambda}{2}\lVert\mathbf{w}\rVert^2 + 
			\frac{1}{\ell}\sum_{i=1}^{\ell}
			\mathcal{L}\big(\mathbf{w},b;(\mathbf{x}_i,\Sigma_i,y_i)\big).
		\end{equation}
		The algorithm receives as input two parameters: (i) the number of iterations, $T$, and (ii) the number of examples to use for calculating sub-gradients, $k$. Initially, we set $\mathbf{w}^{(1)}$ to any vector whose norm is at most $1/\sqrt{\lambda}$ and $b^{(1)}=0$. On the $t$-th iteration, we randomly choose a subset of $\mathcal{X}$, of cardinality $k$, i.e., $\mathcal{X}_t\subseteq\mathcal{X}$, where $\lvert\mathcal{X}_t\rvert=k$, and set the learning rate to $\eta_t=\frac{1}{\lambda t}$. Then, we approximate the objective function of the above optimization problem with
		$$
			\frac{\lambda}{2}\lVert\mathbf{w}\rVert^2 + 
			\frac{1}{k}\sum_{(\mathbf{x}_i,\Sigma_i,y_i)\in\mathcal{X}_t}
			\mathcal{L}\big(\mathbf{w},b;(\mathbf{x}_i,\Sigma_i,y_i)\big).
		$$
		Then, we perform the update steps
		$$
			\mathbf{w}^{(t+1)}\leftarrow\mathbf{w}^{(t)}-\frac{\eta_t}{k}\frac{\partial J}{\partial\mathbf{w}},
			\quad
			b^{(t+1)}\leftarrow b^{(t)}-\frac{\eta_t}{k}\frac{\partial J}{\partial b},
		$$
		where the first-order derivatives are given in (\ref{eq:dJ_dw}), (\ref{eq:dJ_db}), if the training is conducted in the original space (Sect.~\ref{ssec:svmgsu}), or in (\ref{eq:dJprime_dw}), (\ref{eq:dJprime_db}), if the learning is conducted in linear subspaces (Sect.~\ref{ssec:svmgsu_sl}). Last, we project $\mathbf{w}^{(t+1)}$ onto the ball of radius $1/\sqrt{\lambda}$, i.e., the set $\mathcal{B}=\{\mathbf{w}\colon\lVert\mathbf{w}\rVert\leq1/\sqrt{\lambda}\}$. The output of the algorithm is the pair of $\mathbf{w}^{(T+1)}$, $b^{(T+1)}$. Algorithm~\ref{alg:sgd} describes the proposed method in pseudocode.
		  
		\begin{algorithm}
			\caption{A stochastic sub-gradient descent algorithm for solving SVM-GSU.}\label{alg:sgd}
			\begin{algorithmic}[1]
				\Inputs{$\mathcal{X}$, $\lambda$, $T$, $k$}
				\Initialize{$b^{(1)}=0$, $\mathbf{w}^{(1)}$ such that $\lVert\mathbf{w}^{(1)}\rVert\leq\frac{1}{\sqrt{\lambda}}$}
				\For{$t=1,2,\ldots,T$}
				      \State Choose $\mathcal{X}_t\subseteq\mathcal{X}$, where $\lvert\mathcal{X}_t\rvert=k$
				      \State Set $\eta_t=\frac{1}{\lambda t}$
				      \State $\mathbf{w}^{(t+1)}\leftarrow\mathbf{w}^{(t)}-\frac{\eta_t}{k}\frac{\partial J}{\partial\mathbf{w}}$
				      \State $\mathbf{w}^{(t+1)}\leftarrow\min\Big(1,\frac{1/\sqrt{\lambda}}{\lVert\mathbf{w}^{(t+1)}\rVert}\Big)\mathbf{w}^{(t+1)}$
				      \State $b^{(t+1)}\leftarrow b^{(t)}-\frac{\eta_t}{k}\frac{\partial J}{\partial b}$
				\EndFor
			\end{algorithmic}
		\end{algorithm}

\section{Experiments}\label{sec:experiments}

	    
	In this section we first illustrate the workings of the proposed linear SVM-GSU classifier on a synthetic $2$D toy example (Sect.~\ref{ssec:exp_toy}) and then apply the algorithm on five different classification problems using publicly available and popular datasets. Here, we summarize how the uncertainty is modeled in each case, so as to illustrate how our framework can be applied in practice.

	\begin{figure*}[t]
		\centering
		\begin{subfigure}[b]{0.23\linewidth}
			\includegraphics[width=\linewidth]{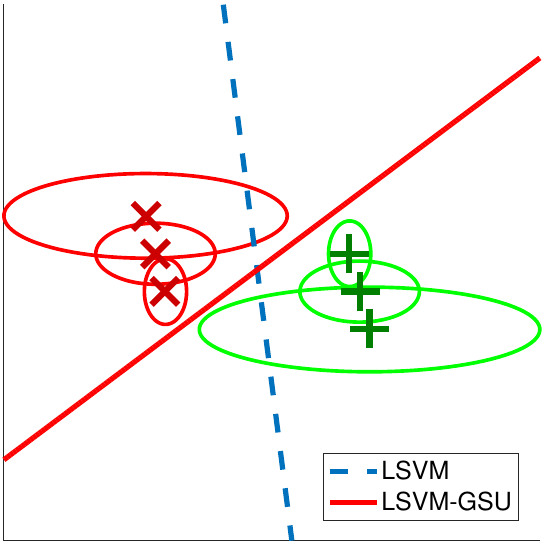}
			\caption{}
			\label{subfig:lsvmgsu_svm}
		\end{subfigure}
		~
		\begin{subfigure}[b]{0.23\linewidth}
			\includegraphics[width=\linewidth]{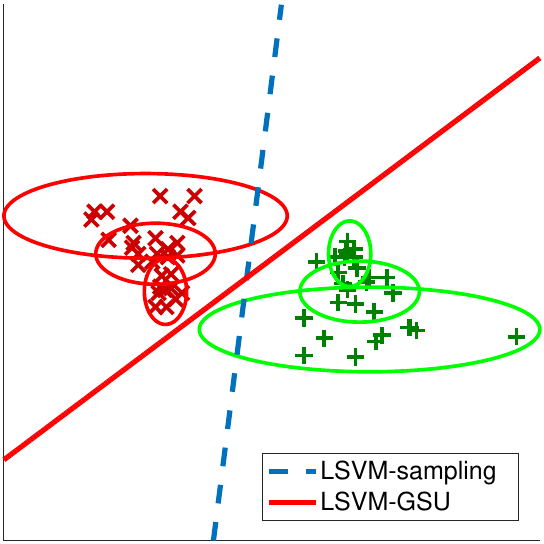}
			\caption{$N=10$}
			\label{subfig:lsvmgsu_svm_sam_10}
		\end{subfigure}
		~
		\begin{subfigure}[b]{0.23\linewidth}
			\includegraphics[width=\linewidth]{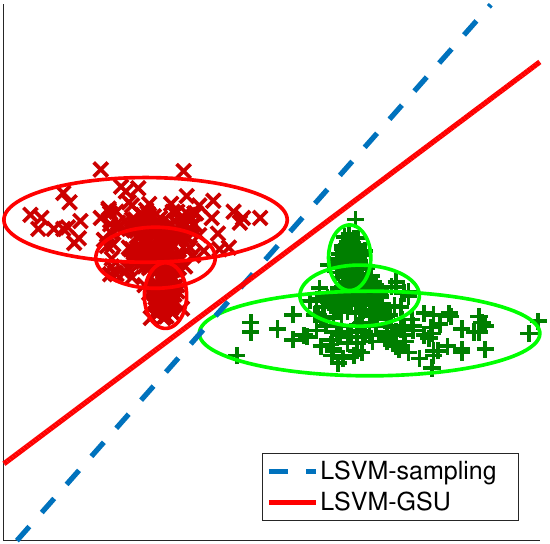}
			\caption{$N=10^2$}
			\label{subfig:lsvmgsu_svm_sam_100}
		\end{subfigure}
		~
		\begin{subfigure}[b]{0.23\linewidth}
			\includegraphics[width=\linewidth]{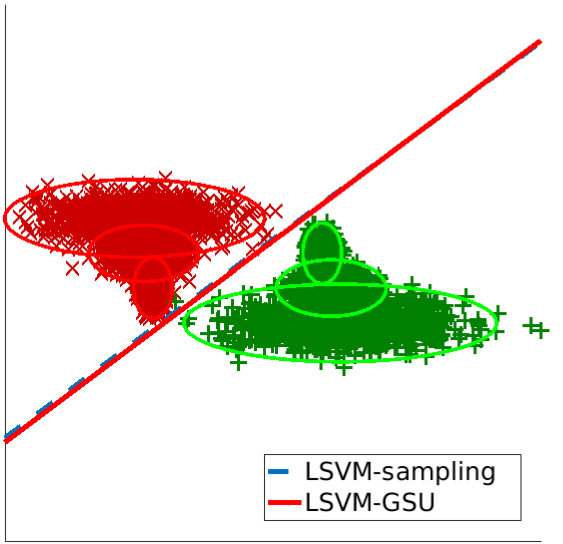}
			\caption{$N=10^3$}
			\label{subfig:lsvmgsu_svm_sam_1000}
		\end{subfigure}
		\caption{Toy example illustrating on $2$D data, (a) the proposed LSVM-GSU (red solid line) in comparison with the standard LSVM (blue dashed line), and (b)-(d) with the standard SVM that learns by sampling from the input Gaussians (LSVM-sampling), where $N$ is the sampling size.}
		\label{fig:toy_ex}
	\end{figure*}
	    
	First, we address the problem of image classification of handwritten digits (Sect.~\ref{ssec:exp_mnist}) using the MNIST dataset. As we show in Appendix~\ref{app:app_B}, by using a first-order Taylor approximation around a certain image with respect to some common image transformations (small translations in our case), we show that the images that would be produced by those translations would follow a Gaussian distribution with mean the image in question and a covariance matrix whose elements are functions of the derivatives of the image intensities/color with respect to those transformations. In the simple case of spatial translations, the covariance elements are functions of the spatial gradients. This is a case where the uncertainty is modeled. We show that our method outperforms the linear SVM and other SVM variants that handle uncertainty isotropically.
	    
	Second, we address the binary classification problem using the Wisconsin Diagnostic Breast Cancer (WDBC) dataset (Sect.~\ref{ssec:exp_wdbc}). This is a case in which each data example summarizes a collection of samples by their second order statistics. More specifically, each data example contains as features the mean and the variance of measurements on several cancer cells -- mean and variances over the different cells. With our formulation we obtain state of the art results on this dataset.
	    
	Third, we address the problem of emotional analysis using electroencephalogram (EEG) signals (Sect.~\ref{ssec:exp_deap}). In this case, we exploit a very popular method for estimating the power spectrum of time signals; namely the Welch method, which allows for estimating not only the mean values of the features (periodograms), but also their variances, making it suitable for using the proposed SVM-GSU.
	    
	Fourth, we address the problem of detection of advertisements in TV news videos (Sect.~\ref{ssec:exp_tvnews}). This is an interesting case where uncertainty information is given only for a few dimensions of the input space, rendering inapplicable the methods that treat uncertainty isotropically. In contrast, the proposed method can model such uncertainty types using low-rank covariance matrices.
	    
	Finally, we address the challenging problem of complex event detection in video (Sect.~\ref{ssec:exp_med}). We used the $\sim$5K outputs of a pre-trained DCNN in order to extract a representation for each frame in a video and calculated the mean and covariances over the frames of a video in order to classify it. This is a second example in which the mean and the covariance matrices are calculated from data. We show that our formulation outperforms the linear SVM and other SVM variants that handle uncertainty isotropically.

	\subsection{Toy example using synthetic data}\label{ssec:exp_toy}
	    
		In this subsection, we present a toy example on $2$D data that provides insights to the way the proposed algorithm works. As shown in Fig.~\ref{subfig:lsvmgsu_svm}, negative examples are denoted by red $\times$ marks, while positive ones by green crosses. We assume that the uncertainty of each training example is given via a covariance matrix. For illustration purposes, we draw the iso-density loci of points at which the value of the PDF of the Gaussian is the $0.03\%$ of its maximum value.

		First, a baseline linear SVM (LSVM) is trained using solely the centres of the distributions; i.e., ignoring the uncertainty of each example. The resulting separating boundary is the dashed blue line in Fig.~\ref{subfig:lsvmgsu_svm}. The proposed linear SVM-GSU (LSVM-GSU) is trained using both the centres of the above distributions and the covariance matrices. The resulting separating boundary is the solid red line in Fig.~\ref{subfig:lsvmgsu_svm}. It is clear that the separating boundaries can be very different and that the solid red line is a better one given the assumed uncertainty modeling.
		  
\begin{figure}[]
			\centering
			\includegraphics[width=0.9\linewidth]{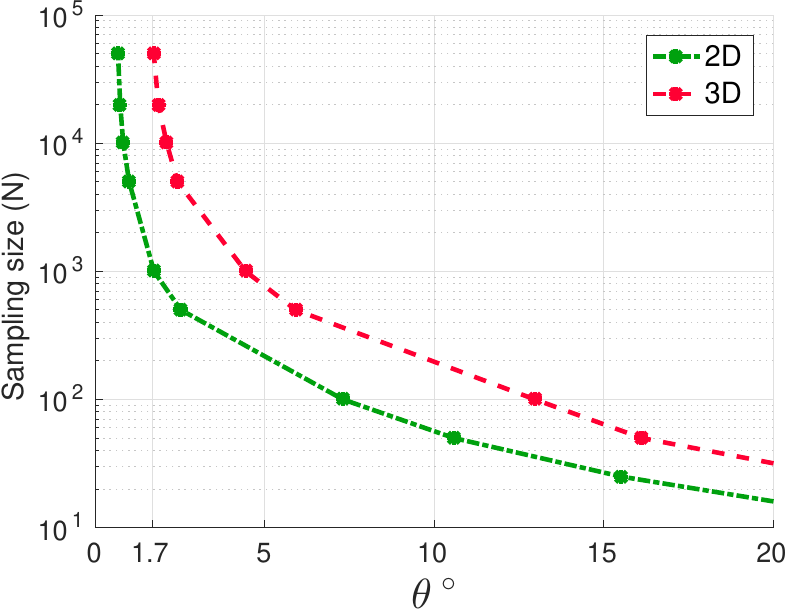}
			\caption{Difference between the separating hyperplanes of LSVM-GSU and the standard LSVM with sampling (angle $\theta$), when varying the number of samples used in the standard SVM, for the $2$D and $3$D toy datasets.}
			\label{fig:thetas}
		\end{figure}

		Next, we investigate on how many samples are needed in order to obtain LSVM-GSU's separating line by sampling $N$ samples from each Gaussian and using the standard LSVM (LSVM-sampling). The results for various values of $N$ are depicted in Fig.~\ref{fig:toy_ex}, where it is clear that ones needs almost $3$ orders of magnitude more examples. In order to investigate how this number changes with the dimensionality of the feature space we performed the same experiment in a similar $3$D dataset. In Fig.~\ref{fig:thetas} we plot the angle between the hyperplanes obtained by the LSVM-GSU and the LSVM-sampling for both the $2$D and the $3$D datasets. We observe that, in the $3$D case, we need at least one order of magnitude more samples from each Gaussian, compared to the $2$D case; that is, in the $2$D case, we obtain $\theta\approx1.7^\circ$ using $N=10^3$ samples from each Gaussian, while in the $3$D case, the sampling size for obtaining the same approximation ($\theta\approx1.7^\circ$) is $N=5\times10^4$. This is indicative of the difficulties of using the sampling approach when dealing with high-dimensional data, where the number of dimensions is in the hundreds or thousands.
		  
	\subsection{Hand-written digit classification}\label{ssec:exp_mnist}
	    
		\subsubsection{Dataset and experimental setup}\label{ssec:exp_mnist_data_setup}
		
			The proposed algorithm is also evaluated in the problem of image classification using the MNIST dataset of handwritten digits~\cite{bottou1994comparison}. The MNIST dataset provides a training set of $60$K examples (approx. $6000$ examples per digit), and a test set of $10$K examples (approx. $1000$ examples per digit). Each sample is represented by a $28\times28$ $8$-bit image.
			
			In order to make the dataset more challenging, as well as to model a realistic distortion that may happen to this kind of images, the original MNIST dataset was ``polluted'' with noise. More specifically, each image example was rotated by a random angle uniformly drawn from $[-\theta, +\theta]$, where $\theta$ is measured in degrees. Moreover, each image was translated by a random vector $\mathbf{t}$ uniformly drawn from $[-t_p,+t_p]^2$, where $t_p$ is a positive integer expressing distance that is measured in pixels. We created five different noisy datasets by setting $\theta=\ang{15}$ and $t_p\in\{3,5,7,9,11\}$, resulting in the polluted datasets $D_1$ to $D_5$, respectively. $D_0$ denotes the original MNIST dataset.
			
			\begin{table*}[!t]
				\caption{MNIST ``$1$'' versus ``$7$'' experimental results in terms of testing accuracy. The proposed LSVM-GSU is compared to the baseline linear SVM (LSVM), Power SVM (PSVM)~\cite{zhang2012power}, and a linear SVM extension which handles the uncertainty isotropically (LSVM-iso), as in \cite{bi2004support, qi2013robust}.}
				\label{tab:mnist_exp_res}
				\begin{tabular}{|c|c|c|c|c|c|c|c|}
				    \hline
				    \multicolumn{2}{|c|}{\textbf{Dataset}}                                                                   & $\mathbf{D_0}$         & $\mathbf{D_1}$         & $\mathbf{D_2}$         & $\mathbf{D_3}$         & $\mathbf{D_4}$         & $\mathbf{D_5}$         \\ \hline\hline
				    \multicolumn{2}{|c|}{LSVM}                                                                               & 0.9952                 & 0.9362                 & 0.8240                 & 0.6830                 & 0.6558                 & 0.6027                 \\ \hline
				    \multicolumn{2}{|c|}{PSVM~\cite{zhang2012power}}                                                         & 0.9963                 & 0.9315                 & 0.8157                 & 0.7017                 & 0.6650                 & 0.6259                 \\ \hline
				    \multicolumn{2}{|c|}{LSVM-iso (as in \cite{bi2004support, qi2013robust})}                                & 0.9968                 & 0.9327                 & 0.8133                 & 0.7222                 & 0.6675                 & 0.6328                 \\ \hline
				    \multirow{2}{*}{\begin{tabular}[c]{@{}c@{}}LSVM-GSU\end{tabular}}         & Learning in original space   & 0.9971                 & 0.9452                 & 0.8310                 & 0.7216                 & 0.6708                 & 0.6353                 \\ \cline{2-8} 
													      & Learning in linear subspaces & \textbf{0.9972} (0.99) & \textbf{0.9480} (0.97) & \textbf{0.8562} (0.89) & \textbf{0.7543} (0.85) & \textbf{0.6974} (0.95) & \textbf{0.6640} (0.25) \\ \hline
			      \end{tabular}
			\end{table*}
			
			We created six different experimental scenarios using the above datasets ($D_0$-$D_5$). First, we defined the problem of discriminating the digit one (``$1$'') from the digit seven (``$7$'') similarly to \cite{ghio2012nested}. Each class in the training procedure consists of $25$ samples, randomly chosen from the pool of digits one ($6k$ totally) and seven ($6k$ totally), while the evaluation of the trained classifier is carried out on the full testing set ($2k$ examples). In each experimental scenario we report the average of $100$ runs and we compare the proposed linear SVM-GSU (LSVM-GSU) to the baseline linear SVM (LSVM), Power SVM~\cite{zhang2012power}, and LSVM-iso (a variation of SVM formulation that handles only isotropic uncertainty, similarly to~\cite{bi2004support, qi2013robust}). We report the testing accuracy and the mean testing accuracy across $100$ runs. Finally, we repeat the above experiments for various sizes of the training set; i.e., using $25$, $50$, $100$, $500$, $1000$, $3000$, $6000$ positive examples per digit, in order to investigate how this affects the results.

			\subsubsection{Uncertainty modeling}\label{ssec:exp_mnist_uncertainty}
			
				In Appendix~\ref{app:app_B}, we propose a methodology that, given an image, models the distribution of the images that result by small random translations of it. We show that under a first-order Taylor approximation of the image intensities/color with respect to those translations, and the assumption that the translations are small and follow a Gaussian distribution, the resulting distribution of the images is also a Gaussian with mean the original image and a covariance matrix whose elements are functions of the image derivatives with respect to the transforms -- in this case functions of the image spatial gradients. The derivation could be straightforwardly extended to other transforms (e.g. rotations, scaling).

				In our experiments in this dataset we set the variances of the horizontal and the vertical components of the translation, denoted by $\sigma_h^2$ and $\sigma_v^2$ respectively, to $\sigma_h^2 = \sigma_v^2 = \left(\frac{p_t}{3}\right)^2$, so that the translation falls in the square $[-p_t,p_t]\times[-p_t,p_t]$ with probability $99.7\%$. The $p_t$ is measured in pixels and for the experiments described below, it is set to $p_t=5$ pixels.

			\subsubsection{Experimental results}\label{ssec:exp_mnist_results}
		  
				Table~\ref{tab:mnist_exp_res} shows the performance of the proposed classifier (LSVM-GSU) and the compared techniques in terms of testing accuracy for each dataset defined above ($D_0$-$D_5$). The optimization of the training parameter for the various SVM variants was performed using a line search on a $3$-fold cross-validation procedure. The performance of LSVM-GSU when the training of each classifier is carried out in the original feature space is shown in row $5$, and in linear subspaces in row $6$. In row $6$ we report both the classification performance, and in parentheses the fraction of variance that resulted in the best classification result.
			
\begin{figure}[t]%
					\centering
					\begin{subfigure}[b]{.45\columnwidth}
						\includegraphics[width=\columnwidth]{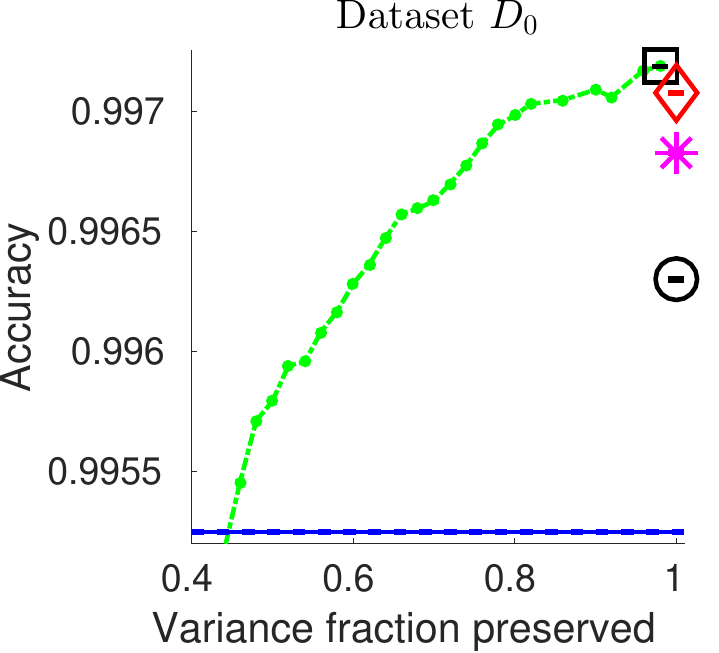}%
						\caption{}
						\label{subfiga}%
					\end{subfigure}
					~
					\begin{subfigure}[b]{.45\columnwidth}
						\includegraphics[width=\columnwidth]{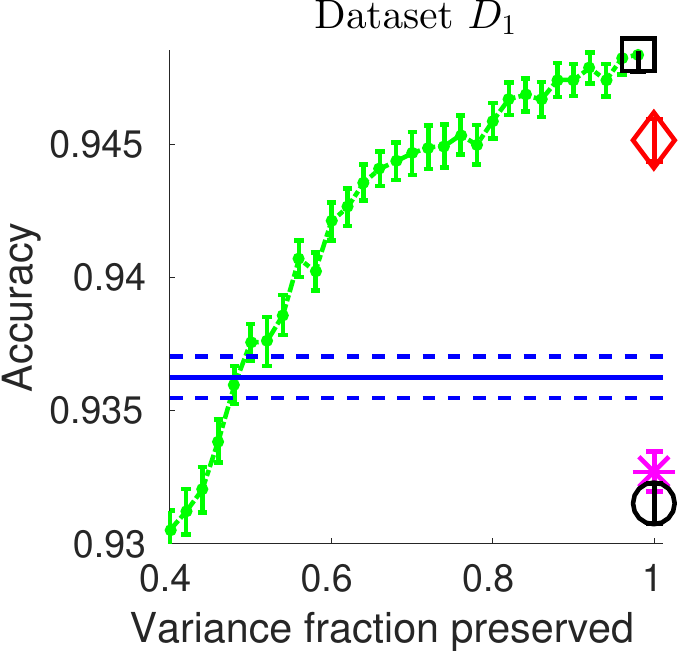}%
						\caption{}%
						\label{subfigb}%
					\end{subfigure}
					\\
					\begin{subfigure}[b]{.45\columnwidth}
						\includegraphics[width=\columnwidth]{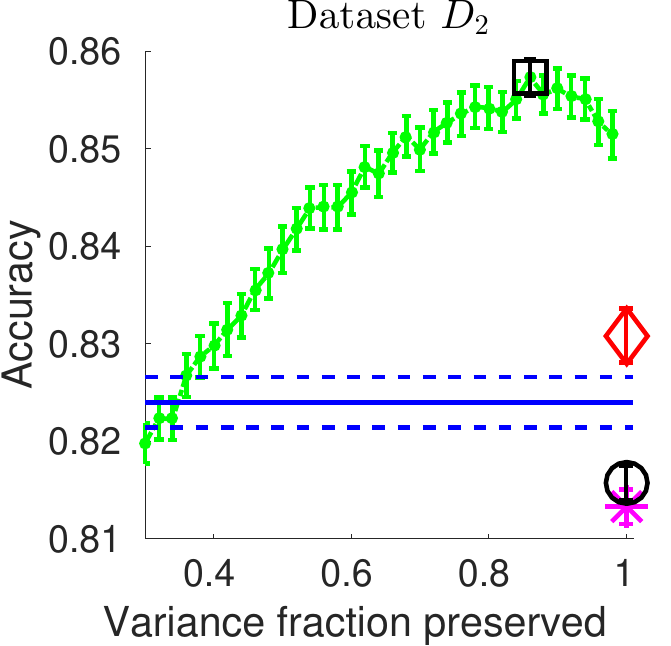}%
						\caption{}%
						\label{subfigc}%
					\end{subfigure}
					~
					\begin{subfigure}[b]{.45\columnwidth}
						\includegraphics[width=\columnwidth]{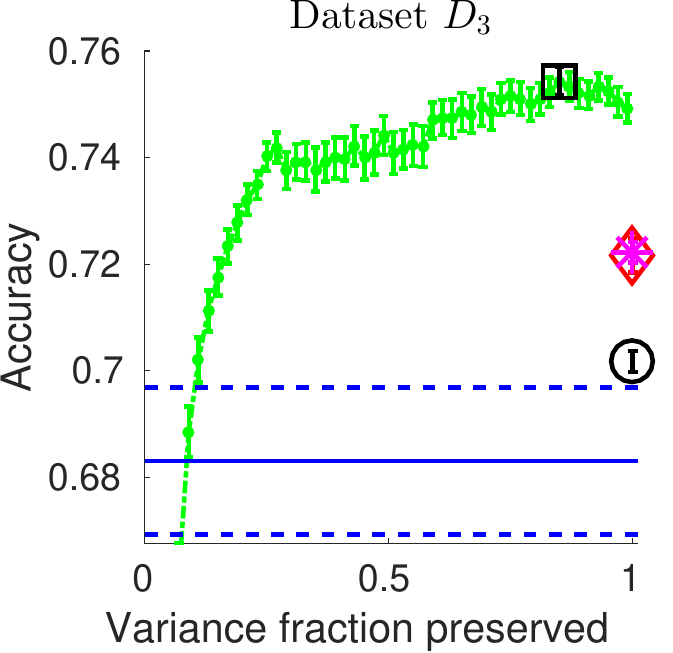}%
						\caption{}%
						\label{subfigd}%
					\end{subfigure}
					\\
					\begin{subfigure}[b]{.45\columnwidth}
						\includegraphics[width=\columnwidth]{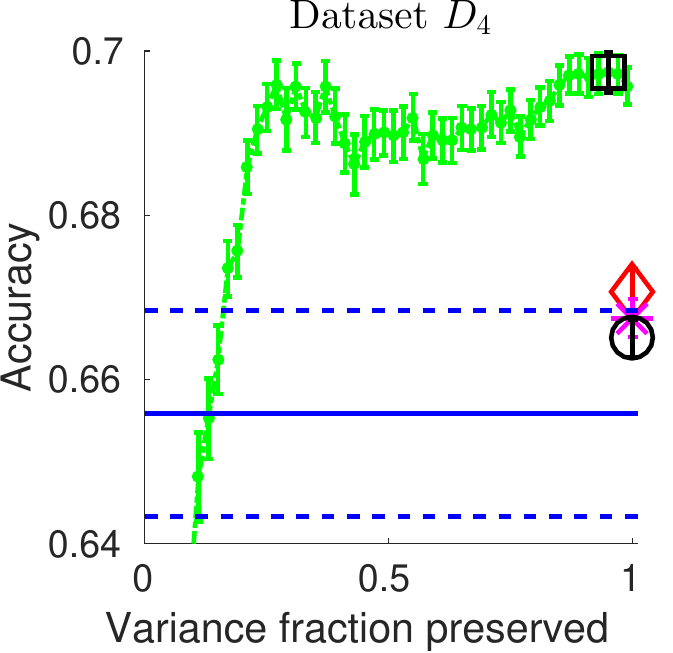}%
						\caption{}%
						\label{subfige}%
					\end{subfigure}
					~
					\begin{subfigure}[b]{.45\columnwidth}
						\includegraphics[width=\columnwidth]{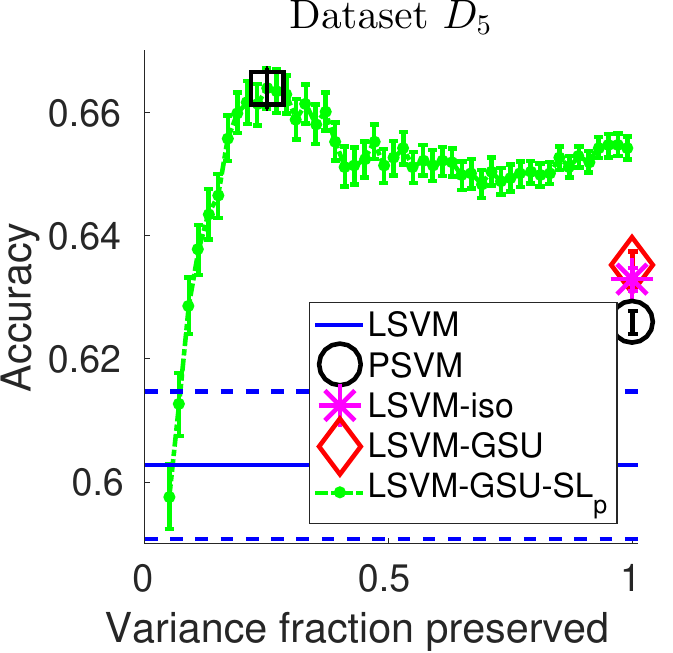}%
						\caption{}%
						\label{subfigf}%
					\end{subfigure}%
					\caption{Comparisons between the proposed LSVM-GSU, the baseline LSVM, and the LSVM with isotropic noise in (a) the original MNIST dataset ($D_0$), and (b)-(f) the noisy generated datasets $D_1$-$D_5$.}
					\label{fig:mnist_res}
				\end{figure}

				The performance of the baseline linear SVM (LSVM) is shown in the second row, the performance of Power SVM (PSVM)~\cite{zhang2012power} is shown in the third row, and the performance of the linear SVM extension, based on the proposed formulation, handling the noise isotropically, as in~\cite{bi2004support, qi2013robust}, (LSVM-iso) is shown in the fourth row. Moreover, Fig.~\ref{fig:mnist_res} shows the results of the above experimental scenarios for datasets $D_0$-$D_5$. The horizontal axis of each subfigure describes the fraction of the total variance  preserved for each covariance matrix, while the vertical axis shows the respective performance of LSVM-GSU with learning in linear subspaces ($\text{LSVM-GSU-SL}_p$). Furthermore, in each subfigure, for $p=1$ we draw the result of LSVM-GSU in the original feature space (denoted with a rhombus), the result of PSVM~\cite{zhang2012power} (denoted with a circle), as well as the result of LSVM-iso~\cite{bi2004support, qi2013robust} (denoted with a star).
				
\begin{figure}[t]
					\centering
					\begin{subfigure}[b]{0.48\linewidth}
						\includegraphics[width=\linewidth]{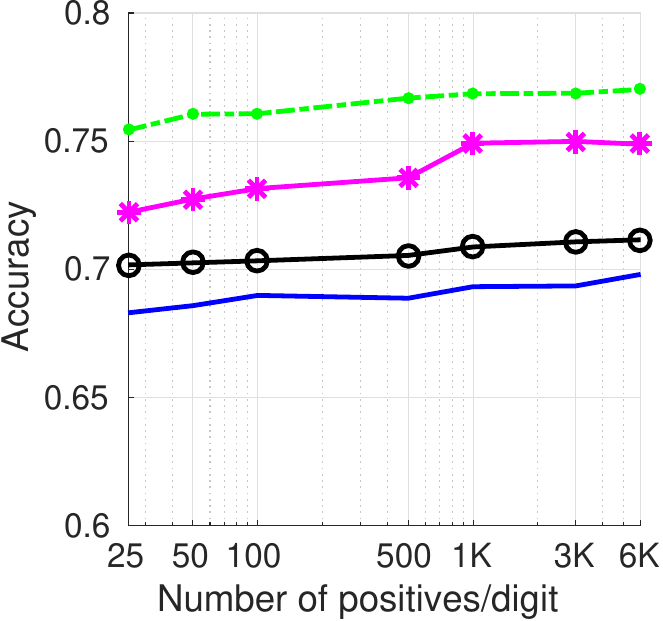}
						\caption{}
						\label{subfig:mnist_d3}
					\end{subfigure}
					~
					\begin{subfigure}[b]{0.48\linewidth}
						\includegraphics[width=\linewidth]{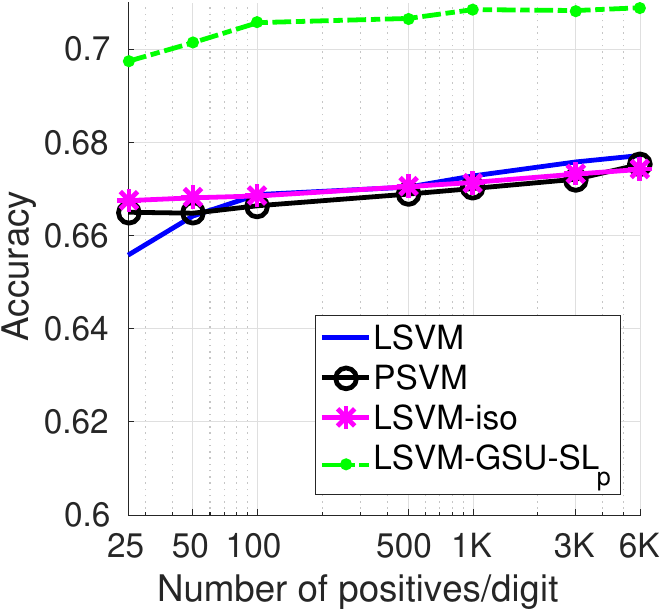}
						\caption{}
						\label{subfig:mnist_d4}
					\end{subfigure}
					\caption{MNIST ``$1$'' versus ``$7$'' experimental results using $25$, $50$, $100$, $500$, $1000$, $3000$, $6000$ positive examples per digit. The proposed LSVM-GSU using learning linear subspaces (LSVM-GSU-S\textsubscript{p}) is compared to the baseline linear SVM (LSVM), Power SVM (PSVM)~\cite{zhang2012power}, and a linear SVM extension which handles the uncertainty isotropically (LSVM-iso), as in \cite{bi2004support, qi2013robust}. The fraction of variance preserved for the proposed method is (a) $p=0.85$ (dataset $D_3$), (b) $p=0.95$ (dataset $D_4$). Very similar results are observed for all other datasets.}
					\label{fig:mnist}
				\end{figure}

				We report the mean, and with an error-bar show the variance of the $100$ iterations. The performance of the baseline LSVM is shown with a solid line, while two dashed lines show the corresponding variance of the $100$ runs. From the obtained results, we observe that the proposed LSVM-GSU with learning in linear subspaces outperforms LSVM, PSVM, and LSVM-iso for all datasets $D_0$-$D_5$. Moreover, LSVM-GSU achieves better classification results than PSVM in all datasets, and than LSVM-iso in $5$ out of $6$ datasets, when learning is carried out in the original feature space. Finally, all the reported results are shown to be statistically significant using the t-test~\cite{hines2008probability}; significance values ($p$-values) were much lower than the significance level of $1\%$. Finally, in Fig.~\ref{fig:mnist}, we show the experimental results using various training set sizes and we observe that this does not qualitatively affect the behavior of the various compared methods.

	\subsection{Wisconsin Diagnostic Breast Cancer dataset}\label{ssec:exp_wdbc}
		  
		The Wisconsin Diagnostic Breast Cancer (WDBC) dataset~\cite{Lichman2013} consists of features computed from $569$ images, each belonging to one of the following two classes: \textit{malignant} (212 instances) and \textit{benign} (357 instances). The digitized images depict breast mass obtained by Fine Needle Aspirate (FNA) and they describe characteristics of the cell nuclei present in the image. Each feature vector is of the form $\mathbf{x}=(x_1,\ldots,x_{10},s_1,\ldots,s_{10},w_1,\ldots,w_{10})^\top\in\mathbb{R}^{30}$, where $x_j$ is the mean value, $s_j$ the standard error, and $w_j$ the largest value of the $j$-th feature, $j=1,\ldots,10$. Ten real-valued features are computed for each cell nucleus.
		  
		Since the standard error $s_i$ and variance $\sigma_i^2$ are connected via the relation $s_i = \frac{\sigma_i^2}{N}$, where $N$ is the (unknown) size of the sample where standard deviation was computed, we assign to each input example a diagonal covariance matrix given by $\Sigma_i=\operatorname{diag}(\sigma_1^2,\ldots,\sigma_{10}^2,\sigma_0^2,\ldots,\sigma_0^2)\in\mathbb{S}^{30}_{++}$, where $\sigma_0^2$ is set to a small positive constant (e.g., $10^{-6}$) indicating very low uncertainty for the respective features, and $\sigma_j^2$ is computed using the standard error by scaling the standard error values into the range of mean values; that is, the maximum variance is set to $80\%$ of the range of the corresponding mean value.
		  
		The proposed algorithm is compared in terms of testing accuracy both to the baseline linear SVM (LSVM), Power SVM~\cite{zhang2012power} (PSVM), and to LSVM-iso, similarly to Sect.~\ref{ssec:exp_mnist}. Since the original dataset does not provide a division in training and evaluation subsets, we divided the dataset randomly into a training subset ($90\%$) and an evaluation subset ($10\%$). The optimization of the $\lambda$ parameter for all classifiers was performed using a line search on a $10$-fold cross-validation procedure. We repeated the experiment $10$ times and report the average results in Table~\ref{tab:wdbc}. The results are statistically significant and show the superiority of LSVM-GSU. More specifically, we used the t-test~\cite{hines2008probability} and obtained significance values ($p$-values) lower than $0.05$.
		  
		\begin{table}[h]
			\centering
			\caption{Comparison between the proposed LSVM-GSU, the baseline LSVM, Power SVM, and LSVM-iso.}
			\label{tab:wdbc}
				\begin{tabular}{|l||c|}
				\hline
				\textbf{Classifier}                                  & \textbf{Testing Accuracy} \\ \hline\hline
				LSVM                                                 &         95.15\%           \\ \hline
				PSVM~\cite{zhang2012power}                           &         96.37\%           \\ \hline
				LSVM-iso (as in~\cite{bi2004support, qi2013robust})  &         96.53\%           \\ \hline
				LSVM-GSU (proposed)                                  & \textbf{97.14}\%          \\ \hline
				\end{tabular}
		\end{table}

	\subsection{Emotion analysis using physiological signals}\label{ssec:exp_deap}
	
		\subsubsection{Dataset and experimental setup}\label{sssec:exp_deap_data_setup}
		  
			For evaluating the proposed method in the domain of emotional analysis using physiological signals, we used the publicly available DEAP~\cite{koelstra2012deap} dataset, which provides EEG features of $32$ participants who were recorded while watching $40$ one-minute long excerpts of music videos. Three different binary classification problems were defined: the classification of low/high arousal, low/high valence and low/high liking videos.
			
			From the EEG signals, power spectral features were extracted using the Welch method~\cite{welch1967use}. The logarithms of the spectral power from \textit{theta} ($4-8$ Hz), \textit{slow alpha} ($8-10$ Hz), \textit{alpha} ($8-12$ Hz), \textit{beta} (12-30Hz), and \textit{gamma} ($30+$ Hz) bands were extracted from all $32$ electrodes as features, similarly to~\cite{koelstra2012deap}. In addition to power spectral features, the difference between the spectral power of all the symmetrical pairs of electrodes on the right and left hemisphere was extracted to measure the possible asymmetry in the brain activities due to emotional stimuli. The total number of EEG features of a video for $32$ electrodes is $216$. For feature selection, we used Fisher's linear discriminant similarly to~\cite{koelstra2012deap}.

		\subsubsection{Uncertainty modeling}\label{sssec:exp_deap_uncertainty}
		  
			For modeling the uncertainty of each training example, we used a well-known property of the Welch method~\cite{welch1967use} for estimating the power spectrum of a time signal. First, the time signal was divided into (overlapping or non-overlapping) windows, where the periodogram was computed for each window. Then the resulting frequency-domain values were averaged over all windows. Besides these mean values, that are the desired outcomes of the Welch method, we also computed the variances, and, thus, each 216-element vector was assigned with a diagonal covariance matrix.

		\subsubsection{Experimental results}\label{sssec:exp_deap_results}
		  
			Table~\ref{tab:deap_res} shows the performance of the proposed linear SVM-GSU (LSVM-GSU) in terms of accuracy and F1 score for each target class in comparison to LSVM, PSVM~\cite{zhang2012power}, and LSVM-iso, similarly to Sect.~\ref{ssec:exp_mnist} and \ref{ssec:exp_wdbc}, as well as the Naive Bayesian (NB) classifier used in~\cite{koelstra2012deap}. For each participant, the F1 measure was used to evaluate the performance of emotion classification in a leave-one-out cross validation scheme. At each step of the cross validation, one video was used as the test-set and the rest were used for training. For optimizing the $\lambda$ parameter of the various SVM classifiers, we used a line search on a $3$-fold cross-validation procedure.
		  
			From the obtained results, we observe that the proposed algorithm achieved better classification performance than LSVM, PSVM, LSVM-iso, as well as the NB classifier used in~\cite{koelstra2012deap} for all three classes, in terms of testing accuracy, and for the two out of three classes in terms of F1 score.
			
			\begin{table}
				\centering
				\caption{Comparisons between the proposed LSVM-GSU, the baseline NB, LSVM, Power SVM, and the LSVM with isotropic noise.}
				\label{tab:deap_res}
				\begin{tabular}{|l|p{0.5cm}|p{0.5cm}|p{0.5cm}|p{0.5cm}|p{0.5cm}|p{0.5cm}|}
					\cline{2-7}
					\multicolumn{1}{c|}{}                                                       & \multicolumn{2}{c||}{\textbf{Arousal}} & \multicolumn{2}{c||}{\textbf{Valence}} & \multicolumn{2}{c|}{\textbf{Liking}} \\ \hline
					\multicolumn{1}{|l|}{\textbf{Classifier}}                                   & \textbf{ACC}      & \textbf{F1}        & \textbf{ACC}      & \textbf{F1}        & \textbf{ACC}     & \textbf{F1}       \\ \hline
					\multicolumn{1}{|l|}{\textbf{NB~\cite{koelstra2012deap}}}                   & 0.620             & \textbf{0.583}     & 0.576             & 0.563              & 0.554            & 0.502             \\ \hline
					\multicolumn{1}{|l|}{\textbf{LSVM}}                                         & 0.626             & 0.451              & 0.616             & 0.538              & 0.655            & 0.470             \\ \hline
					\multicolumn{1}{|l|}{\textbf{PSVM~\cite{zhang2012power}}}                   & 0.625             & 0.521              & 0.633             & 0.561              & 0.651            & 0.522             \\ \hline
					\multicolumn{1}{|l|}{\textbf{LSVM-iso~\cite{bi2004support, qi2013robust}}}  & 0.645             & 0.531              & 0.645             & 0.603              & 0.658            & 0.530             \\ \hline
					\multicolumn{1}{|l|}{\textbf{LSVM-GSU}}                                     & \textbf{0.659}    & 0.551              & \textbf{0.650}    & \textbf{0.609}     & \textbf{0.666}   & \textbf{0.539}    \\ \hline
				\end{tabular}
			\end{table}

	\subsection{TV News Channel Commercial Detection}\label{ssec:exp_tvnews}
		
		\subsubsection{Dataset and experimental setup}\label{sssec:exp_tvnews_data_setup}
		  
			The proposed algorithm is evaluated in the problem of detection of advertisements in TV news videos using the publicly available and very large dataset of~\cite{vyas2014commercial}. This dataset comprises $120$ hours of TV news broadcasts from CNN, CNNIBN, NDTV, and TIMES NOW (approximately $22k$, $33k$, $17k$, and $39k$ videos, respectively). The authors of~\cite{vyas2014commercial} used various low-level audio and static-, motion-, and text-based visual features, to extract and provide a $4125$-dimensional representation for each video, that includes the variance values for $24$ of the above features. For a detailed description of the dataset, see~\cite{vyas2014commercial}. 
		  
		\subsubsection{Uncertainty modeling}\label{sssec:exp_tvnews_uncertainty}
		  
			This dataset represents a real-world case where uncertainty information is given only for a few dimensions of the feature space. In this case we model the covariance matrix of each input example as a low-rank diagonal matrix, whose non-zero variance values correspond to the dimensions for which uncertainty is provided. Each such matrix corresponds to a Gaussian with non-zero variance along the few specific given dimensions. Since the information about the input variance is provided just for the $24$ of the $4125$ features, there is no natural way of estimating a single variance value, i.e., an isotropic covariance matrix, for each training example.
		  
		\subsubsection{Experimental results}\label{sssec:exp_tvnews_results}
			
			Table~\ref{tab:tv_comm_det} shows the performance of the proposed linear SVM-GSU (LSVM-GSU) in terms of F1 score in comparison to LSVM, similarly to~\cite{vyas2014commercial}. As discussed above, since methods that model the uncertainty isotropically (such as~\cite{bi2004support, qi2013robust, zhang2012power}), are not applicable in this dataset, we experimented on this dataset using only the proposed algorithm and the standard linear SVM. Following the protocol of~\cite{vyas2014commercial}, we did cross-dataset training and testing. For optimizing the $\lambda$ parameter of both LSVM and LSVM-GSU we used a line search on a $3$-fold cross-validation procedure. From the obtained results, we observe that the proposed algorithm achieved considerably better classification than LSVM in almost all cases (more than $10\%$ relative boost on average).
			
			\begin{table*}[!t]
			      \centering
			      \caption{Comparisons between the proposed LSVM-GSU and the baseline LSVM, similarly to~\cite{vyas2014commercial}.}
			      \label{tab:tv_comm_det}
			      \begin{tabular}{l|l|c|c||c|c||c|c||c|c|}
				    \cline{3-10}
				    \multicolumn{2}{c|}{\multirow{3}{*}{}}                        	          & \multicolumn{8}{c|}{Training on}                                                                                                                            \\ \cline{3-10} 
				    \multicolumn{2}{c|}{}                                                                                   & \multicolumn{2}{c||}{CNN}   & \multicolumn{2}{c||}{CNNIBN} & \multicolumn{2}{c||}{NDTV}        & \multicolumn{2}{c|}{TIMES NOW} \\ \cline{3-10} 
				    \multicolumn{2}{c|}{}                                                                                   & LSVM      & LSVM-GSU        & LSVM     & LSVM-GSU          & LSVM            & LSVM-GSU        & LSVM      & LSVM-GSU           \\ \hline
				    \multicolumn{1}{|l|}{\multirow{4}{*}{\rotatebox[origin=c]{90}{Testing on}}} & CNN                       & 0.7799    & \textbf{0.9589} & 0.7799   & \textbf{0.8050}   & 0.7799          & \textbf{0.8113} & 0.7799    & \textbf{0.9226}    \\ \cline{2-10} 
				    \multicolumn{1}{|l|}{}                            			  & CNNIBN                    & 0.7915    & \textbf{0.8836} & 0.7915   & \textbf{0.9215}   & 0.7915          & \textbf{0.8978} & 0.7915    & \textbf{0.8611}    \\ \cline{2-10} 
				    \multicolumn{1}{|l|}{}                            			  & NDTV                      & 0.8484    & \textbf{0.9248} & 0.8484   & \textbf{0.8565}   & 0.8484          & \textbf{0.9709} & 0.8484    & \textbf{0.8823}    \\ \cline{2-10} 
				    \multicolumn{1}{|l|}{}                            			  & TIMES NOW                 & 0.7809    & \textbf{0.9461} & 0.7809   & \textbf{0.7863}   & \textbf{0.7809} & 0.7493          & 0.7809    & \textbf{0.9421}    \\ \hline
			      \end{tabular}
			\end{table*}

	\subsection{Video Event Detection}\label{ssec:exp_med}
		
		\subsubsection{Dataset and experimental setup}\label{ssec:exp_med_data_setup}
		
			In our experiments on video event detection we used datasets from the challenging TRECVID Multimedia Event Detection (MED) task~\cite{2015trecvidover}. For training, we used the MED 2015 training dataset consisting of the ``pre-specified'' (PS) video subset ($2000$ videos, $80$ hours) and the ``event background'' (Event-BG) video subset ($5000$ videos, $200$ hours). For testing, we used the large-scale ``MED14Test'' dataset~\cite{2015trecvidover, jiang2015bridging} ($\sim24$K videos, $850$ hours). Each video in the above datasets belongs to, either one of $20$ target event classes, or to the ``rest of the world'' (background) class. More specifically, in the training set, $100$ positive and $5000$ negative samples are available for each event class, while the evaluation set includes only a small number of positive (e.g., only $16$ positives for event E021, and $28$ for E031) and approximately $24$K negative videos.
			
			For video representation, approximately $2$ keyframes per second were extracted from each video. Each keyframe was represented using the last hidden layer of a pre-trained deep convolutional neural network (DCNN). More specifically, a $22$-layer inception style network, trained according to the GoogLeNet architecture~\cite{Szegedy15}, was used. This network had been trained on various selections of the ImageNet ``Fall 2011'' dataset and provides scores for $5055$ concepts~\cite{ILSVRC15}.

		\subsubsection{Uncertainty modeling}\label{ssec:exp_med_uncertainty}
			
			Let us now define a set $\mathcal{X}$ of $\ell$ annotated random vectors representing the aforementioned video-level feature vectors. Each random vector is assumed to be distributed normally; i.e., for the random vector representing the $i$-th video, $\mathbf{X}_i$, we have $\mathbf{X}_i\sim\mathcal{N}(\mathbf{x}_i, \Sigma_i)$. That is, $\mathcal{X}=\{(\mathbf{x}_i,\Sigma_i,y_i)\colon\mathbf{x}_i\in\mathbb{R}^{n},\Sigma_i\in\mathbb{S}_{++}^n,y_i\in\{\pm1\},\:i=1,\ldots,\ell\}$. For each random vector $\mathbf{X}_i$, a number, $N_i$, of observations, $\{\mathbf{x}_i^t\in\mathbb{R}^n\colon t=1,\ldots,N_i\}$ are available (these are the keyframe-level vectors that have been computed). Then, the sample mean vector and the sample covariance matrix of $\mathbf{X}_i$ are computed. However, the number of observations per each video that are available for our dataset is in most cases much lower than the dimensionality of the input space. Consequently, the covariance matrices that arise are typically low-rank; i.e. $\operatorname{rank}(\Sigma_i)\leq N_i\leq n$. To overcome this issue, we assumed that the desired covariance matrices are diagonal. That is, we require that the covariance matrix of the $i$-th training example is given by $\widehat{\Sigma_i} = \operatorname{diag}\big(\hat{\sigma}_i^1,\ldots,\hat{\sigma}_i^n\big)$, such that the squared Frobenious norm of the difference $\Sigma_i-\widehat{\Sigma_i}$ is minimum. That is, the estimator covariance matrix $\widehat{\Sigma_i}$ must be equal to the diagonal part of the sample covariance matrix $\Sigma_i$, i.e. $\widehat{\Sigma_i} = \operatorname{diag}\big(\sigma_i^1,\ldots,\sigma_i^n\big)$. We note that, using this approximation approach, the covariance matrices are diagonal but anisotropic and different for each training input example. This is in contrast with other methods (e.g. \cite{zhang2012power, bi2004support, qi2013robust}) that assume more restrictive modeling for the uncertainty; e.g., isotropic noise for each training sample.

		\subsubsection{Experimental results}\label{ssec:exp_med_results}

			\begin{table*}[!t]
			\centering
			\caption{Event detection performance (AP and MAP) of the linear SVM-GSU compared to the baseline linear SVM, Power SVM~\cite{zhang2012power}, and a LSVM extension for handling isotropic uncertainty (as in \cite{bi2004support, qi2013robust}) using the MED15 (for training) and MED14Test (for testing) datasets.}
			\label{tab:med_res}
			      \begin{tabular}{|c|c|l|c|c|c|c|l|c|c|c|}
			      \hline
			      \multirow{2}{*}{\textbf{\begin{tabular}[c]{@{}c@{}}Event\\ Class\end{tabular}}} & \multicolumn{5}{c|}{\textbf{\begin{tabular}[c]{@{}c@{}}Feature Configuration 1\\ (5055-D)\end{tabular}}}                                                                                                                                                  & \multicolumn{5}{c|}{\textbf{\begin{tabular}[c]{@{}c@{}}Feature Configuration 2\\ (10110-D)\end{tabular}}}                                                                                                                                                 \\ \cline{2-11} 
													      & \textbf{LSVM}    & \textbf{\begin{tabular}[c]{@{}l@{}}PSVM\\ ~\cite{zhang2012power}\end{tabular}} & \textbf{\begin{tabular}[c]{@{}c@{}}LSVM-iso\\ \cite{bi2004support, qi2013robust}\end{tabular}} & \textbf{\begin{tabular}[c]{@{}c@{}}LSVM-GSU\\ (proposed)\end{tabular}} & \textbf{\begin{tabular}[c]{@{}c@{}}McNemar\\ Tests\end{tabular}} & \textbf{LSVM}   & \textbf{\begin{tabular}[c]{@{}l@{}}PSVM\\ ~\cite{zhang2012power}\end{tabular}}  & \textbf{\begin{tabular}[c]{@{}c@{}}LSVM-iso\\ \cite{bi2004support, qi2013robust}\end{tabular}} & \textbf{\begin{tabular}[c]{@{}c@{}}LSVM-GSU\\ (proposed)\end{tabular}} & \textbf{\begin{tabular}[c]{@{}c@{}}McNemar\\ Tests\end{tabular}} \\ \hline
			      E021          & 0.0483           &  0.0510  & 0.0500           & \textbf{0.0515}  & $\ast,\diamond,\sim$  & 0.0829          &  0.0834  & \textbf{0.1074}  & 0.0778           & $\diamond,\sim$       \\ \hline
			      E022          & 0.0227           &  0.0310  & \textbf{0.0350}  & 0.0277           & $\ast,\diamond,\sim$  & 0.0674          &  0.0773  & 0.1023           & \textbf{0.1429}  & $\ast,\diamond,\sim$  \\ \hline
			      E023          & 0.4159           &  0.4515  & \textbf{0.6059}  & 0.6057           & $\ast,\diamond$       & 0.7050          &  0.7236  & 0.7802           & \textbf{0.7943}  & $\ast,\diamond,\sim$  \\ \hline
			      E024          & 0.0071           &  0.0081  & 0.0097           & \textbf{0.0105}  & $\diamond$            & 0.0187          &  0.0223  & \textbf{0.0394}  & 0.0367           & $\ast$                \\ \hline
			      E025          & 0.0052           &  0.0052  & \textbf{0.0074}  & 0.0068           &                       & \textbf{0.0219} &  0.0245  & 0.0161           & 0.0135           & $\diamond$            \\ \hline
			      E026          & 0.0457           &  0.0459  & 0.0606           & \textbf{0.0608}  & $\diamond$            & 0.0731          &  0.0745  & 0.0976           & \textbf{0.1109}  & $\ast,\diamond,\sim$  \\ \hline
			      E027          & \textbf{0.1319}  &  0.1424  & 0.1174           & 0.1219           & $\ast,\diamond,\sim$  & 0.1152          &  0.0133  & 0.1254           & \textbf{0.1812}  & $\ast,\diamond,\sim$  \\ \hline
			      E028          & 0.4242           &  0.4125  & 0.3819           & \textbf{0.4335}  & $\ast,\diamond,\sim$  & 0.1863          &  0.2214  & \textbf{0.2700}  & 0.2278           & $\ast,\diamond,\sim$  \\ \hline
			      E029          & 0.0812           &  0.0914  & \textbf{0.1793}  & 0.1791           & $\diamond$            & 0.2046          &  0.1987  & \textbf{0.2149}  & 0.1999           & $\ast,\diamond,\sim$  \\ \hline
			      E030          & 0.0516           &  0.0551  & 0.0877           & \textbf{0.0884}  &                       & 0.1001          &  0.1276  & 0.1596           & \textbf{0.1774}  & $\ast,\diamond,\sim$  \\ \hline
			      E031          & 0.4416           &  0.4425  & 0.4480           & \textbf{0.4796}  & $\ast,\diamond,\sim$  & 0.7595          &  0.7599  & 0.7422           & \textbf{0.7697}  & $\ast,\diamond,\sim$  \\ \hline
			      E032          & 0.0280           &  0.0400  & 0.0870           & \textbf{0.1196}  & $\ast,\diamond,\sim$  & 0.0989          &  0.1011  & 0.1290           & \textbf{0.1292}  & $\ast,\diamond$       \\ \hline
			      E033          & 0.3483           &  0.3614  & 0.3901           & \textbf{0.4187}  & $\ast,\sim$           & 0.4571          &  0.4789  & 0.5091           & \textbf{0.5164}  & $\ast$                \\ \hline
			      E034          & 0.0583           &  0.0588  & 0.0599           & \textbf{0.0614}  & $\diamond$            & 0.3207          &  0.3214  & 0.3200           & \textbf{0.3380}  & $\ast,\diamond,\sim$  \\ \hline
			      E035          & 0.3330           &  0.3419  & \textbf{0.3500}  & 0.3369           & $\ast,\diamond,\sim$  & \textbf{0.3516} &  0.3419  & 0.3252           & 0.3059           & $\ast,\diamond$       \\ \hline
			      E036          & \textbf{0.0894}  &  0.0748  & 0.0695           & 0.0704           & $\diamond$            & 0.1156          &  0.1186  & 0.1064           & \textbf{0.1288}  & $\ast,\diamond,\sim$  \\ \hline
			      E037          & 0.0884           &  0.0880  & \textbf{0.1981}  & 0.1968           & $\ast,\diamond,\sim$  & 0.1169          &  0.1257  & 0.1598           & \textbf{0.1629}  & $\ast,\diamond,\sim$  \\ \hline
			      E038          & 0.0261           &  0.0241  & 0.0212           & \textbf{0.0291}  & $\diamond$            & \textbf{0.0558} &  0.0498  & 0.0557           & 0.0539           & $\diamond$            \\ \hline
			      E039          & 0.2677           &  0.2698  & \textbf{0.2959}  & 0.2757           & $\ast,\diamond,\sim$  & 0.4188          &  0.4219  & \textbf{0.4349}  & 0.4271           & $\ast,\diamond,\sim$  \\ \hline
			      E040          & 0.0421           &  0.0315  & 0.0375           & \textbf{0.0377}  & $\ast,\diamond$       & 0.0837          &  0.0889  & 0.0856           & \textbf{0.0902}  & $\ast,\diamond$       \\ \hline\hline
			      \textbf{MAP}  & 0.1478           &  0.1513  & 0.1746           & \textbf{0.1806}  & --                    & 0.2177          &  0.2187  & 0.2390           & \textbf{0.2442}  & --                    \\ \hline
			      \end{tabular}
			\end{table*}

			We experimented using two different feature configurations. First, we used the mean vectors and covariance matrices as computed using the method discussed above (Sect.~\ref{ssec:exp_med_uncertainty}). Furthermore, in order to investigate the role of variances in learning with baseline LSVM, we constructed mean vectors and covariance matrices as shown in Table~\ref{tab:feat_config}, where $\sigma_0$ is typically set to a small positive constant (e.g., $10^{-6}$) indicating very low uncertainty for the respective features.
			
			\begin{table}[]
			\centering
			\caption{Mean vector and covariance matrix of the $i$-th example for feature configurations 1 and 2 of the video event detection experiments.}
			\label{tab:feat_config}
			      \begin{tabular}{|c||c|}
				    \hline
				    \multirow{2}{*}{\textbf{Configuration 1}} & $\mathbf{x}_i = (x_{i,1},\ldots,x_{i,n})^\top\in\mathbb{R}^{n}$                                                        \\
										      & $\Sigma_i,= \operatorname{diag}\big(\sigma_i^1,\ldots,\sigma_i^n\big)\in\mathbb{S}_{++}^{n}$                           \\ \hline
				    \multirow{2}{*}{\textbf{Configuration 2}} & $\mathbf{x}_i = (x_{i,1},\ldots,x_{i,n},\sigma_i^1,\ldots,\sigma_i^n)^\top\in\mathbb{R}^{2n}$                          \\
										      & $\Sigma_i,= \operatorname{diag}\big(\sigma_i^1,\ldots,\sigma_i^n,\sigma_0,\ldots,\sigma_0\big)\in\mathbb{S}_{++}^{2n}$ \\ \hline
			      \end{tabular}
			\end{table}
			
			For both feature configurations, Table~\ref{tab:med_res} shows the performance of the proposed linear SVM-GSU (LSVM-GSU) in terms of average precision (AP)~\cite{2015trecvidover, tzelepis2013improving} for each target event in comparison with LSVM, PSVM~\cite{zhang2012power}, and LSVM-iso approaches. Moreover, for each dataset, the mean average precision (MAP) across all target events is reported. The optimization of the $\lambda$ parameter for the various SVMs was performed using a line search on a $10$-fold cross-validation procedure. The bold-faced numbers indicate the best result achieved for each event class. We also report the results of the McNemar~\cite{mcnemar1947note, gkalelis2013mixture}, statistical significance tests. A $\ast$ denotes statistically significant differences between the proposed LSVM-GSU and baseline LSVM, a $\diamond$ denotes statistically significant differences between LSVM-GSU and PSVM, and a $\sim$ denotes statistically significant differences between LSVM-GSU and LSVM-iso.
			
			From the obtained results, we observe that the proposed algorithm achieved better detection performance than LSVM, PSVM, and LSVM-iso, in both feature configurations. For feature configuration 1, the proposed LSVM-GSU achieved a relative boost of $22.2\%$ compared to the baseline standard LSVM and $19.4\%$ compared to Power SVM, while for feature configuration 2 respective relative boosts of $12.7\%$ and $11.7\%$, respectively, in terms of MAP. We also experimented using directly the samples from which the covariance matrix of each example was estimated and obtained inferior results; that is, a MAP of $10.15\%$, compared to LSVM's $14.78\%$ and $18.06\%$ of the proposed SVM-GSU.

\section{Conclusion}\label{sec:conclusion}

	In this paper we proposed a novel classifier that efficiently exploits uncertainty in its input under the SVM paradigm. The proposed SVM-GSU was evaluated on synthetic data and on five publicly available datasets; namely, the MNIST dataset of handwritten digits, the WDBC, the DEAP for emotion analysis, the TV News Commercial Detection dataset and TRECVID MED for the problem of video event detection. For each of the above datasets and problems, either uncertainty information (e.g., variance for each example and for all or some of the input space dimensions) was part of the original dataset, or a method for modeling and estimating the uncertainty of each training example was proposed. As shown in the experiments, SVM-GSU efficiently takes input uncertainty into consideration and achieves better detection or classification performance than standard SVM, previous SVM extensions that model uncertainty isotropically, and other state of the art methods. Finally, we plan to investigate the kernalization of the proposed algorithm and the extensions of it for the problem of regression under Gaussian input uncertainty.

\appendices

\section{On Gaussian-like integrals over halfspaces}\label{app:app_A}

	\begin{theorem}\label{thm:thm_1}
		Let $\mathbf{X}\in\mathbb{R}^n$ be a random vector that follows the multivariate Gaussian distribution with mean vector $\bg{\mu}\in\mathbb{R}^n$ and covariance matrix $\Sigma\in\mathbb{S}_{++}^n$, where $\mathbb{S}_{++}^n$ denotes the space of $n\times n$ symmetric positive definite matrices with real entries. The probability density function of this distribution is given by $f_{\mathbf{X}}\colon\mathbb{R}^n\to\mathbb{R}$,  $f_{\mathbf{X}}(\mathbf{x}) = \frac{1}{(2\pi)^{\frac{n}{2}}|\Sigma|^{\frac{1}{2}}}\exp\left(-\frac{1}{2}(\mathbf{x}-\bg{\mu})^\top\Sigma^{-1}(\mathbf{x}-\bg{\mu})\right)$. Moreover, let $\mathcal{H}$ be the hyperplane given by $\mathbf{a}^\top\mathbf{x}+b=0$. $\mathcal{H}$ divides the Euclidean $n$-dimensional space into two halfspaces, i.e., $\Omega_{\pm}=\{\mathbf{x}\in\mathbb{R}^n\colon\mathbf{a}^\top\mathbf{x}+b\gtrless0\}$, so that $\Omega_{+}\cup\Omega_{-}=\mathbb{R}^n$ and $\Omega_{+}\cap\Omega_{-}=\varnothing$. Then, the integrals $I_{\pm}\colon\mathbb{R}^n\times\mathbb{R}\to\mathbb{R}$, defined as 
		$$
			I_{\pm}(\mathbf{a},b)
			\triangleq
			\int_{\Omega_{\pm}}\!(\mathbf{a}^\top\mathbf{x}+b)f_{\mathbf{X}}(\mathbf{x})\,\mathrm{d}\mathbf{x},
		$$
		are given by
		\begin{multline}\label{eq:I_+_result}
			I_{\pm}(\mathbf{a},b)=
			\frac{d_\mu}{2}
			\left[
			1\pm\operatorname{erf}\left(\frac{d_\mu}{d_\Sigma}\right)
			\right]
			\pm
			\frac{d_\Sigma}{2\sqrt{\pi}}
			\exp\left(-\frac{d_\mu^2}{d_\Sigma^2}\right),
		\end{multline}
		where $d_\mu=\mathbf{a}^\top\bg{\mu}+b$ and $d_\Sigma=\sqrt{2\mathbf{a}^\top\Sigma\mathbf{a}}$.
	\end{theorem}
	    
	\begin{proof}
		We begin with the integral $I_+$. In our approach we will need several coordinate transforms. First, we start with a translation in order to get rid of the mean, $\mathbf{x}=\mathbf{y}+\bg{\mu}$. Then
		\begin{eqnarray}
			I_+(\mathbf{a},b) \nonumber
			= \\ \nonumber
			\frac{1}{(2\pi)^{\frac{n}{2}}|\Sigma|^{\frac{1}{2}}}
			\int_{\Omega_1^+}
			\!
			(\mathbf{a}^\top\mathbf{y}+\mathbf{a}^\top\bg{\mu}+b)
			\exp
			\left(
			-\frac{1}{2} \mathbf{y}^\top \Sigma^{-1} \mathbf{y}
			\right)
			\,
			\mathrm{d}\mathbf{y},
		\end{eqnarray}
		where $\Omega_1^+=\{\mathbf{y}\in\mathbb{R}^n\colon\mathbf{a}^\top\mathbf{y}+\mathbf{a}^\top\bg{\mu}+b\geq0\}$. Next, since $\Sigma\in\mathbb{S}_{++}^n$, there exist an orthonormal matrix $U$ and a diagonal matrix $D$ with positive elements, i.e. the eigenvalues of $\Sigma$, such that $\Sigma=U^\top DU$. Thus, it holds that $\Sigma^{-1}=(U^\top DU)^{-1}=U^{-1}D^{-1}(U^\top)^{-1}=U^\top D^{-1}U$. Then, by letting $\mathbf{z}=U\mathbf{y}$ and $\mathbf{a}_1=U\mathbf{a}$, we have $\mathbf{a}^\top\mathbf{y}=\mathbf{a}^\top(U^{-1}U)\mathbf{y}=\mathbf{a}^\top U^\top U \mathbf{z}=\mathbf{a}_1^\top\mathbf{z}$, and $\mathbf{y}^\top \Sigma^{-1} \mathbf{y} \nonumber = \mathbf{y}^\top (U^\top DU)^{-1} \mathbf{y} = (\mathbf{y}^\top U^\top)D^{-1}(U\mathbf{y}) = (U\mathbf{y})^\top D^{-1}(U\mathbf{y}) = \mathbf{z}^\top D^{-1}\mathbf{z}$. Then
		\begin{eqnarray}
			I_+(\mathbf{a},b) \nonumber
			= \\ \nonumber
			\frac{1}{(2\pi)^{\frac{n}{2}}|\Sigma|^{\frac{1}{2}}}
			\int_{\Omega_2^+}
			\!
			(\mathbf{a}_1^\top\mathbf{z}+\mathbf{a}^\top\bg{\mu}+b)
			\exp
			\left(
			-\frac{1}{2} \mathbf{z}^\top D^{-1} \mathbf{z}
			\right)
			\,
			\mathrm{d}\mathbf{z}, \nonumber
		\end{eqnarray}
		where $\Omega_2^+=\{\mathbf{z}\in\mathbb{R}^n\colon\mathbf{a}_1^\top\mathbf{z}+\mathbf{a}^\top\bg{\mu}+b\geq0\}$, since for the Jacobian $J=\lvert U \rvert$, it holds that $\lvert J \rvert=1$. Now, in order to do rescaling, we set $\mathbf{z}=D^{\frac{1}{2}}\mathbf{v}$ and $\mathbf{a}_2=D^{\frac{1}{2}}\mathbf{a}_1$. Thus, 
		$$
			\mathbf{z}^\top D^{-1}\mathbf{z}
			=
			(D^{\frac{1}{2}}\mathbf{v})^\top D^{-1} (D^{\frac{1}{2}}\mathbf{v})
			=
			\mathbf{v}^\top  ( D^{\frac{1}{2}} D^{-1} D^{\frac{1}{2}} ) \mathbf{v}
			=
			\mathbf{v}^\top\mathbf{v}.
		$$
		Moreover, $\mathbf{a}_1^\top\mathbf{z}=\mathbf{a}_1^\top(D^{\frac{1}{2}}\mathbf{v})=(D^{\frac{1}{2}}\mathbf{a}_1)^\top\mathbf{v}=\mathbf{a}_2^\top\mathbf{v}.$ Also, it holds that $\lvert D\rvert^{\frac{1}{2}}=\lvert\Sigma\rvert^{\frac{1}{2}}$ and $\mathrm{d}\mathbf{z}=\lvert D^{\frac{1}{2}}\rvert\mathrm{d}\mathbf{v}=\lvert\Sigma\rvert^{\frac{1}{2}}\mathrm{d}\mathbf{v}$. Consequently,
		\begin{eqnarray}
			I_+(\mathbf{a},b)
			= 
			\frac{1}{(2\pi)^{\frac{n}{2}}}
			\int_{\Omega_3^+}
			\!
			(\mathbf{a}_2^\top\mathbf{v}+\mathbf{a}^\top\bg{\mu}+b)
			\exp
			\left(
			-\frac{1}{2} \mathbf{v}^\top\mathbf{v}
			\right)
			\,
			\mathrm{d}\mathbf{v}, \nonumber
		\end{eqnarray}
		where $\Omega_3^+=\{\mathbf{v}\in\mathbb{R}^n\colon\mathbf{a}_2^\top\mathbf{v}+\mathbf{a}^\top\bg{\mu}+b\geq0\}$. Let $B$ be an orthogonal matrix such that $B\mathbf{a}_2=\lVert\mathbf{a}_2\rVert\mathbf{e}_n$, which also means that $\mathbf{a}_2=B^\top\lVert\mathbf{a}_2\rVert\mathbf{e}_n$. Moreover, let $\mathbf{m}=B\mathbf{v}$. Then, 
		$\mathbf{a}_2^\top\mathbf{v}=(B^\top\lVert\mathbf{a}_2\rVert\mathbf{e}_n)^\top\mathbf{v} = \lVert\mathbf{a}_2\rVert\mathbf{e}_n^\top(B\mathbf{v}) = \lVert\mathbf{a}_2\rVert\mathbf{e}_n^\top\mathbf{m}$. Moreover, $\mathbf{v}^\top\mathbf{v} = \mathbf{v}^\top(B^{-1}B)\mathbf{v} = \mathbf{m}^\top\mathbf{m}$. Then
		\begin{eqnarray}
			I_+(\mathbf{a},b)
			= 
			\frac{1}{\sqrt{2\pi}}
			\int_c^{+\infty}
			\!
			\big(\lVert\mathbf{a}_2\rVert t + \mathbf{a}^\top\bg{\mu}+b\big)
			\exp
			\left(
			-\frac{1}{2} t^2
			\right)
			\,
			\mathrm{d}t, \nonumber
		\end{eqnarray}
		where $c=-\frac{\mathbf{a}^\top\bg{\mu}+b}{\lVert\mathbf{a}_2\rVert}$. Since $\lVert\mathbf{a}_2\rVert^2 = \mathbf{a}^\top\Sigma\mathbf{a}$,
		\begin{equation*}
			I_+(\mathbf{a},b)
			= 
			\frac{1}{\sqrt{2\pi}}
			\int_c^{+\infty}
			\!
			\big(\sqrt{\mathbf{a}^\top\Sigma\mathbf{a}} t + \mathbf{a}^\top\bg{\mu}+b\big)
			\exp
			\left(
			-\frac{1}{2} t^2
			\right)
			\,
			\mathrm{d}t,
		\end{equation*}
		which is easily evaluated as~(\ref{eq:I_+_result}). Following similar arguments as above, we arrive at $I_-$.
	\end{proof}

\section{Modeling the uncertainty of an image}\label{app:app_B}

	Let $\mathbf{f}(\mathbf{0})=(f_1(\mathbf{0}),\ldots,f_j(\mathbf{0}),\ldots,f_n(\mathbf{0}))^\top\in\mathbb{R}^n$ be an image with $n$ pixels in row-wise form, and let $\mathbf{f}(\mathbf{t})=(f_1(\mathbf{t}),\ldots,f_j(\mathbf{t}),\ldots,f_n(\mathbf{t}))^\top\in\mathbb{R}^n$ be a translated version of it by $\mathbf{t}=(h,v)^\top$ pixels. Clearly, $f_j\colon\mathbb{R}^2\to\mathbb{R}$ denotes the intensity function of the $j$-th pixel, after a translation by $\mathbf{t}$.

	We will use the multivariate Taylor's theorem in order to approximate the intensity function of the $j$-th pixel of the given image; i.e., function $f_j$. That is, the intensity is approximated as $f_j(\mathbf{t}) = f_j(\mathbf{0}) + \nabla^\top f_j(\mathbf{0})\mathbf{t}$. Then,
	\begin{equation}
		\mathbf{f}(\mathbf{t}) = \mathbf{f}(\mathbf{0}) +
		\begin{pmatrix}
			\nabla^\top f_1(\mathbf{0}) \\
			\vdots                      \\
			\nabla^\top f_n(\mathbf{0}) \\
		\end{pmatrix}
		\mathbf{t}.
	\end{equation}
	Let us now assume that $\mathbf{t}$ is a random vector distributed normally with mean $\bg{\mu}_t$ and covariance matrix $\Sigma_t$, i.e. $\mathbf{t}\sim\mathcal{N}(\bg{\mu}_t,\Sigma_t)$. Then, $\mathbf{X}=\mathbf{f}(\mathbf{t})$ is also distributed normally with mean vector and covariance matrix that are given, respectively, by
	\begin{equation}\label{eq:X_i_mean}
		\bg{\mu}_X
		= 
		\mathbf{f}(\mathbf{0}) +
		\begin{pmatrix}
			\nabla^\top f_1(\mathbf{0}) \\
			\vdots                      \\
			\nabla^\top f_n(\mathbf{0}) \\
		\end{pmatrix}
		\mathbb{E}\left[\mathbf{t}\right],
	\end{equation}
	and
	\begin{equation}\label{eq:X_i_cov}
		\Sigma_X
		=
		\begin{pmatrix}
			\nabla^\top f_1(\mathbf{0}) \\
			\vdots                      \\
			\nabla^\top f_n(\mathbf{0}) \\
		\end{pmatrix}
		\Sigma_t 
		\begin{pmatrix}
			\nabla^\top f_1(\mathbf{0}) \\
			\vdots                      \\
			\nabla^\top f_n(\mathbf{0}) \\
		\end{pmatrix}^\top.
	\end{equation}
	Thus, if $\mathbf{t}\sim\mathcal{N}(\bg{\mu}_t,\Sigma_t)$, then $\mathbf{X}\sim\mathcal{N}(\bg{\mu}_X,\Sigma_X)$, where the mean vector $\bg{\mu}_X$ and the covariance matrix $\Sigma_X$ are given by (\ref{eq:X_i_mean}) and (\ref{eq:X_i_cov}), respectively.

\section*{Acknowledgment}

	This work was supported by the EU's Horizon 2020 programme H2020-693092 MOVING. We would also like to thank the authors of~\cite{sarafis2015building} for providing an implementation of Power SVM.

\ifCLASSOPTIONcaptionsoff
\newpage
\fi

\bibliographystyle{IEEEtran}
\bibliography{svmgsu_pami2017}


\vfill
\vspace{-1cm}

\begin{IEEEbiography}[{\includegraphics[width=1in,height=1.25in,clip,keepaspectratio]{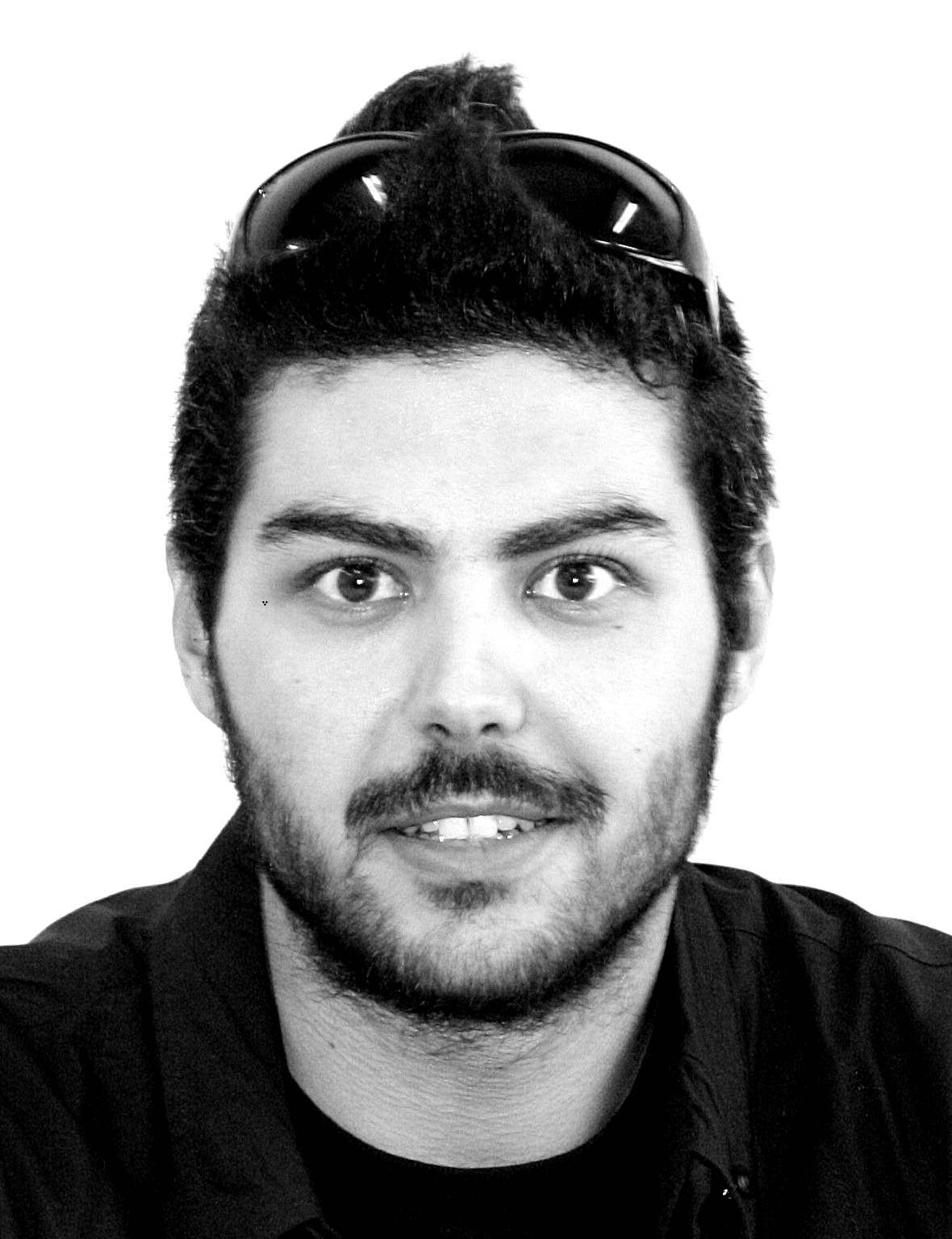}}]{Christos Tzelepis}
received the Diploma degree in electrical engineering from Aristotle University of Thessaloniki, Greece, in 2011. During his diploma thesis, he focused on machine learning techniques with training data of variable reliability. Currently, he is a PhD student in Electronic Engineering and Computer Science at Queen Mary, University of London, within the field of discriminative machine learning, and with ITI/CERTH.
\end{IEEEbiography}

\vfill
\vspace{-1cm}

\begin{IEEEbiography}[{\includegraphics[width=1in,height=1.25in,clip,keepaspectratio]{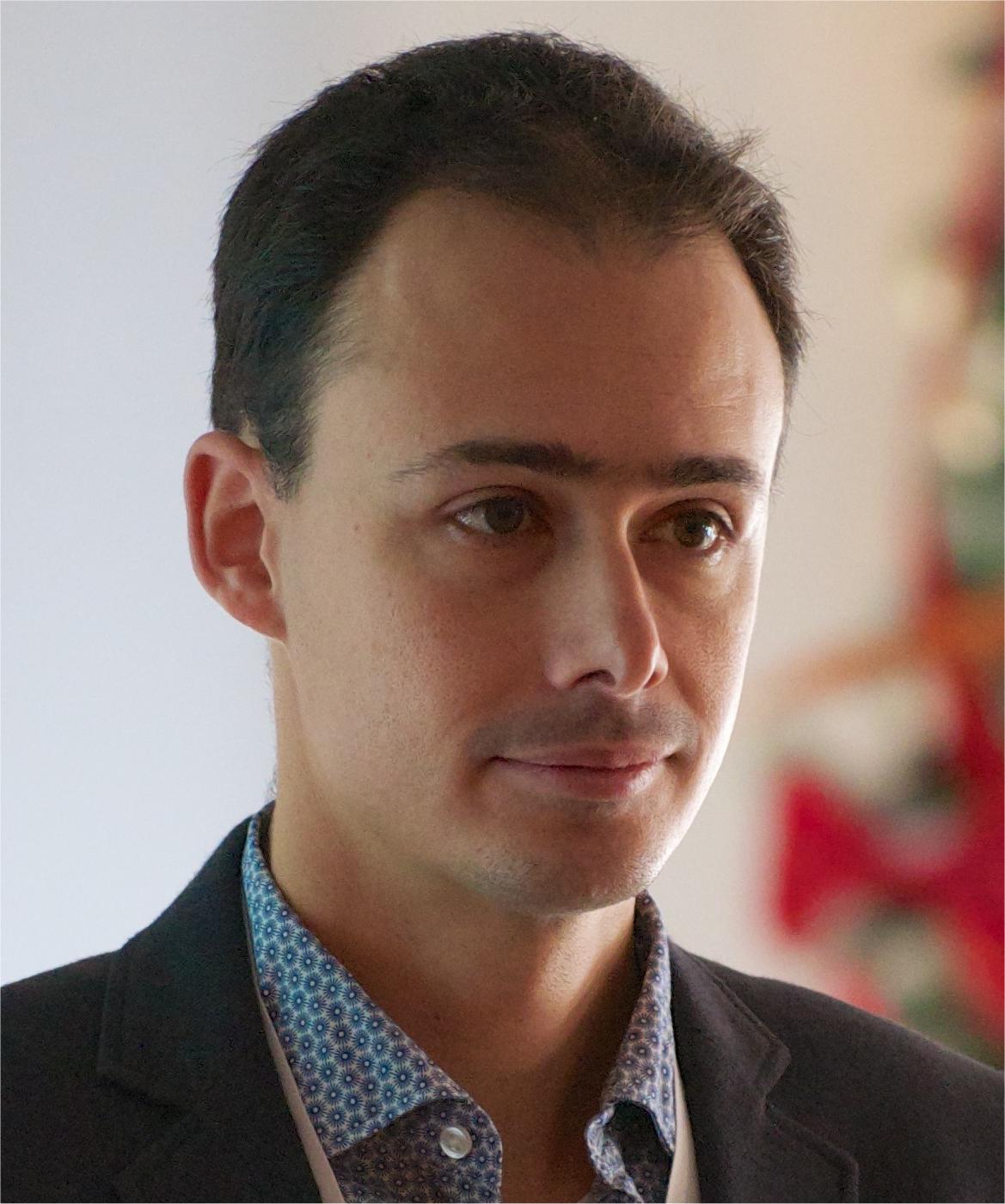}}]{Vasileios Mezaris}
received the BSc and PhD in Electrical and Computer Engineering from the Aristotle University of Thessaloniki in 2001 and 2005, respectively. He is a Senior Researcher (Researcher B) at the Information Technologies Institute (ITI) of the Centre for Research of Technology Hellas (CERTH). His research interests include image and video analysis, event detection in multimedia, machine learning for multimedia, and image and video retrieval. He has co-authored more than 35 journal papers, 10 book chapters, 140 conference papers and 3 patents. He is/was an Associate Editor for the IEEE Signal Processing Letters (2016-present) and IEEE Tran. on Multimedia (2012-2015), and is a Senior Member of the IEEE.
\end{IEEEbiography}

\vfill
\vspace{-1cm}

\begin{IEEEbiography}[{\includegraphics[width=1in,height=1.25in,clip,keepaspectratio]{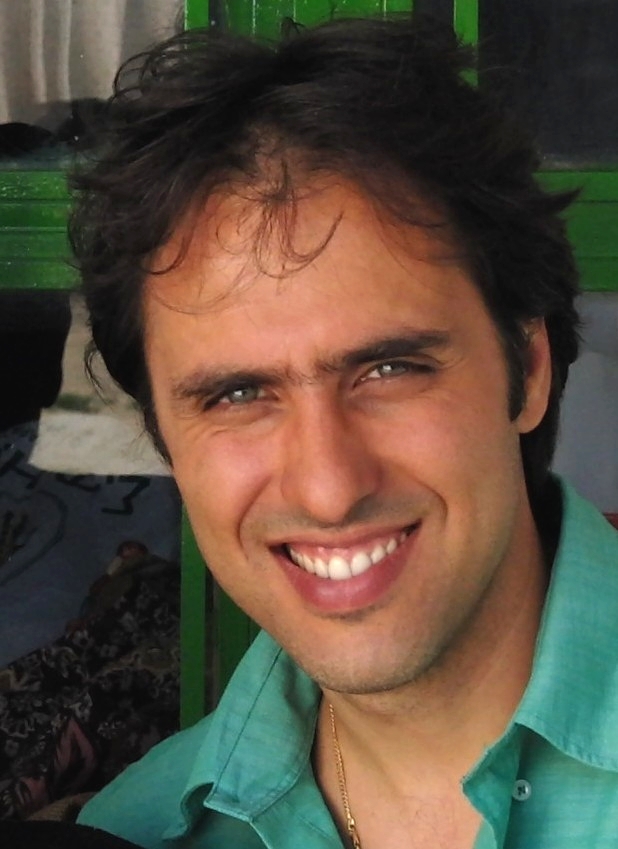}}]{Ioannis Patras}
received the BSc and MSc degrees in computer science from the Computer Science Department, University of Crete, Heraklion, Greece, in 1994 and 1997, respectively, and the PhD degree from the Department of Electrical Engineering, Delft University of Technology, Delft (TU Delft), The Netherlands, in 2001. He is a Reader (Associate Professor) in the School of Electronic Engineering and Computer Science Queen Mary University of London, London, U.K. His current research interests are in the area of Computer Vision, Machine Learning and Affective Computing, with emphasis on the analysis of visual data depicting humans and their activities. He is an Associate Editor of the Image and Vision Computing Journal, Pattern Recognition, and Computer Vision and Image Understanding.
\end{IEEEbiography}

\vfill


\end{document}